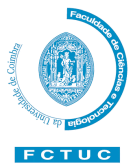

FACULTY OF SCIENCES
AND TECHNOLOGY
UNIVERSITY OF COIMBRA

FCTUC

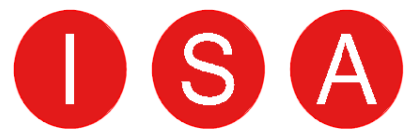

# BW-Eye

# Ophthalmologic Decision Support System based on Clinical Workflow and Data Mining Techniques – Image Registration Algorithm

**Sistema de Apoio à Decisão para a especialidade de Oftalmologia baseado em técnicas de Workflow e Data Mining aplicadas em dados obtidos em ambiente clínico – Algoritmo de Registo de Imagem**

## Project Report
Version 1.0

Ricardo Filipe Alves Martins – Nb. 2002125451

September 8, 2008

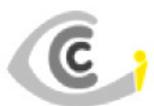

Centro Cirúrgico de Coimbra

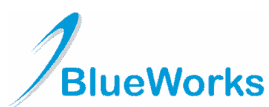

BlueWorks- Medical Expert Diagnosis, Lda

# Document Revisions

| Date | Version | Description | Author |
|------|---------|-------------|--------|
| 04-08-2008 to 04-09-2008 | 1.0 | -Elaboration of the document | Ricardo Martins |

# Acknowledgments

I want to thank Engª Armanda Santos, Engº Edgar Ferreira, Engº Paulo Barbeiro, Liliana Ferraz and Rui Maricato for the good ambient at *Blueworks* and the suggestions, critics and support that they gave me. I also want to thank the management board of *Blueworks* for the opportunity given.

Most important of all, I want to thank my parents, brother, family and friends for the support they gave me whenever I needed and the encouragement in the moments I were frustrated, sad, disappointed and unmotivated.

A special thank to all the people that I met in the last six years that did not treat me neither like a machine nor a robot, but like a normal person that has the same difficulties, problems and capacities of the other persons.

# Abstract

*Blueworks – Medical Expert Diagnosis* is developing an application, *BW-Eye*, to be used as an ophthalmology consultation decision support system. The implementation of this application involves several different tasks and one of them is the implementation of an ophthalmology images registration algorithm.

The work reported in this document is related with the implementation of an algorithm to register images of angiography, colour retinography and red-free retinography. The implementations described were developed in the software *MATLAB*.

The implemented algorithm is based in the detection of the bifurcation points (y-features) of the vascular structures of the retina that usually are visible in the referred type of images. There are proposed two approaches to establish an initial set of features correspondences. The first approach is based in the maximization of the mutual information of the bifurcation regions of the features of images. The second approach is based in the characterization of each bifurcation point and in the minimization of the Euclidean distance between the descriptions of the features of the images in the descriptors space. The final set of the matching features for a pair of images is defined through the application of the RANSAC algorithm.

Although, it was not achieved the implementation of a full functional algorithm, there were made several analysis that can be important to future improvement of the current implementation.

# Resumo

A empresa Blueworks-Medical Expert Diagnosis está a desenvolver uma aplicação, BW-Eye, em que se pretende implementar um sistema de apoio à decisão médica para a especialidade de oftalmologia. O desenvolvimento desta aplicação abrange diversas actividades sendo uma delas a implementação de um algoritmo de registo de imagens de modalidades tipicamente utilizadas em oftalmologia.

Neste documento descrevem-se as tarefas realizadas, relacionadas com a implementação de um algoritmo de registo de imagens de angiografia, retinografia red-free e retinografia a cores. A implementação descrita neste documento foi efectuada utilizando-se o software Matlab.

O algoritmo implementado baseia-se na detecção de pontos de bifurcação das estruturas vasculares do olho que são visíveis através dos tipos de imagens referidos anteriormente. Propõem-se duas abordagens para o estabelecimento de um conjunto inicial de correspondências entre os pontos de bifurcação detectados em pares de imagens. Numas das abordagens o conjunto inicial de correspondências é estabelecido através da maximização da informação mútua das regiões de bifurcação de pares de pontos de bifurcação das imagens a registar. Na segunda abordagem proposta, procede-se à caracterização de cada um dos pontos de bifurcação detectados. O conjunto inicial de correspondências entre pontos de bifurcação de duas imagens é estabelecido pela minimização da distância euclidiana entre a descrição de pares de pontos. Em ambas as abordagens o conjunto final de correspondências entre pares de imagens é definido utilizando-se o algoritmo RANSAC.

Embora não tenha sido atingida uma implementação completamente funcional do algoritmo de registo de imagem, foram feitas certas análises que podem ser importantes para melhorar o desempenho do algoritmo já implementado.

**Palavras-chave:**
registo de imagem, processamento de imagem, oftalmologia, y-features, retina

# Table of Contents

# Figures Index

# Tables Index

# Diagrams Index

# Definitions and Acronyms

**Table 1 – Definitions and acronyms**

| Acronym | Description |
|---------|-------------|
| CT | Computed Tomography |
| fMRI | Functional Magnetic Resonance Imaging |
| MRI | Magnetic Resonance Imaging |
| OCT | Optical Coherence Tomography |
| PET | Positron Emission Tomography |
| RANSAC | Random Sample Consensus |
| SPECT | Single Positron Emission Tomography |

# 1. Introduction

## *1.1 Overview*

Nowadays, as consequence of a big variety of medical instrumentation and techniques, the result of a set of exams of a patient can include text, numbers, graphics images of different modalities, video, sound, oral speech and others. To analyze them, the doctor must have the ability to conjugate the information extracted from each one to produce a diagnosis. Sometimes that information additionally needs to be related and correlated with information from the clinical history and personal data of the patient. One problem that may arise in the practice of medicine is the subjectivity in the extraction of information from the data resulting from each exam and in the conjugation of that information. Thus, the subjectivity in the practice of medicine may cause different doctors to extract different diagnosis from the same set of exams and available data.

Since its creation, *Blueworks* began to develop a project called *BW-Eye Ophthalmologic Decision Support System* which wants to introduce concepts and technology from engineering, maths, physics into the medical activities and in the organization of all processes and tasks implemented in a clinical/hospital institution. The introduction of those principles has an impact in the activities of all professionals (physicians, nurses, auxiliary staff and management) and must be faced as a tool to improve the quality of the service provided to the patients and the effectiveness and productivity of the staff. The project *BW-Eye* consists in the implementation of a decision support system for ophthalmology. It will be a powerful tool that wants to innovate the way how diagnoses are done and to make diagnosis more rigorous and fast. The research is focused in the extraction of useful information from the clinical data available and in the comparison of the clinical case in study with others clinical cases already analyzed and diagnosed by a doctor. The review of the diagnosis, applied therapies and the outcomes obtained in older clinical cases, will also be useful in order to help the doctor in the decision making process. This will be possible due to the combination of the human knowledge provided by the staff of the *Centro Cirúrgico de Coimbra* - it is available a great amount of clinical data that results from the daily activity of the clinic that will be useful - and the data processing capacities of the computers, artificial intelligence science and data mining techniques.

In order to implement *BW-Eye* and to foresee the impact it will have in the traditional clinical organization and planning of tasks, there were made several studies described in [1] [2] [3], both about the internal organization and services provided by *Centro Cirúrgico de Coimbra* and the software and applications already installed.

### 1.1.1 Workflow Analysis

Although in [1] [2] [3] it was studied the workflow involved in the typical examination of a patient for ophthalmology consultation in a specific clinic, the analysis made and the obtained results can be extrapolated to other clinics worldwide and to other consultation specialties.

In result of those studies, it was possible to describe the current workflow of the ophthalmology consultation service of the *Centro Cirúrgico de Coimbra*.

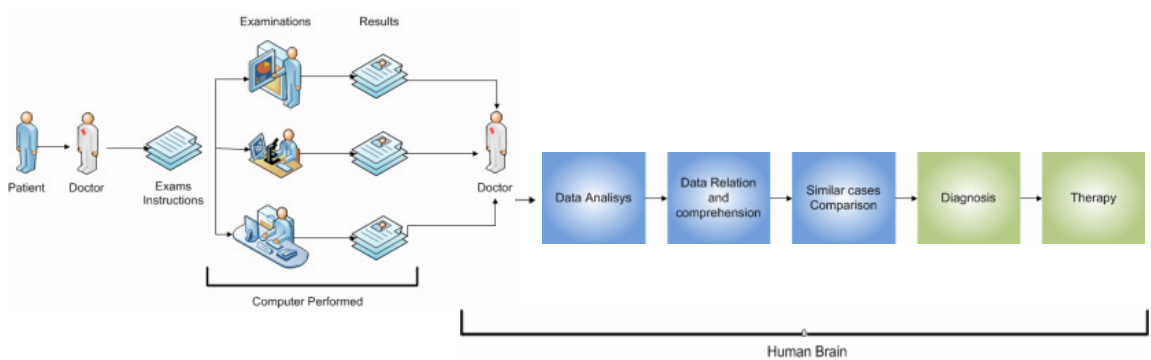

**Diagram 1 - Ophthalmology consultation service in *Centro Cirúrgico de Coimbra*: current workflow (adapted from an internal document of *Blueworks*)**

Typically, when a patient arrives at clinic, he is examined by a doctor who may suggest the realization of some exams, if required. The patient goes to the examination department where a technician carry out the exams required by the doctor. The patient returns to the office of the doctor with the results of the exams and then the doctor proceeds to the analysis of the results and indicate a diagnosis and therapy.

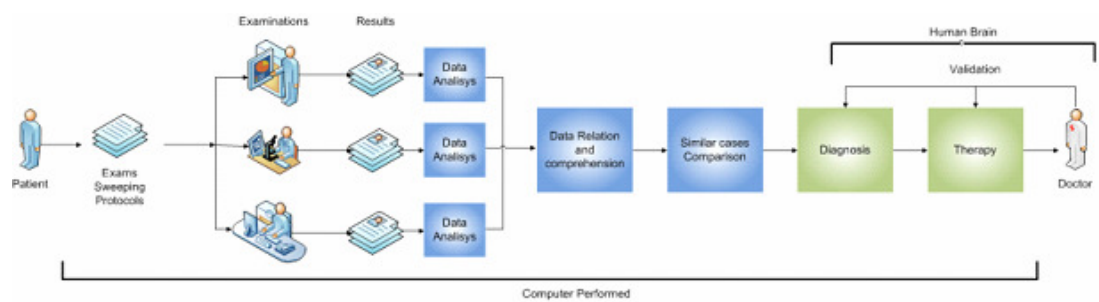

**Diagram 2 - Ophthalmology consultation service in *Centro Cirúrgico de Coimbra*: ideal workflow after the implementation of *BW-Eye* (adapted from an internal document of *Blueworks*)**

With the introduction of the application being developed, it is expected that some alterations occur in the traditional workflow of the consultation of *Cento Cirúrgico de Coimbra*. When a patient arrives, he is submitted to some pre determined protocol in order to the *BW-Eye* suggest a group of exams to be done by the patient, according to the main symptoms identified. In a first phase of the implementation of *BW-Eye* after the pre-determined examinations, the patient goes to the office of the doctor that analysis the results and if required orders more exams to be carried out. After the doctor has all the information needed, it is indicated a diagnose and therapy. In a complete implementation of *BW-Eye* the results of the pre-determined exams will be automatically processed and useful information extracted. The information extracted and additional information from the clinical history of the patient, other patient clinical cases already correctly validated, are correlated and compared in order to identify some similarity between those cases and the case of the patient. The final result of the processing task may be a suggestion of diagnosis and therapy or the necessity to do additional exams to obtain more information. The doctor responsible for that patient will validate the diagnosis and therapy suggested. The cases correctly validated and not validated are added to the database of *BW-Eye*. With the introduction of *BW-Eye* it is expected that the diagnosis of a patient can be obtained faster than usual and can be more reliable and objective.

## 1.1.2 Application Architecture

One of the biggest challenges in the implementation of *BW-Eye* is the necessity to collect the great amount of data produced by the clinical tasks in an organized and structured mode. Thus, it is required that the clinical institution will be equipped with an extensive and robust computer network, a big data storage capacity and the associated backup system. Apart from the infrastructure referred, it is also required an information system to collect the data produced in the clinic (exams devices, people, another information systems), structure and store them.

In [1] [2] [3] it is reported a survey made at *Centro Cirúrgico de Coimbra* where were identified the equipments used to do the exams usually required by ophthalmologists, the clinical cases in which they are required, the type of data and information they provide and the interface available for the integration of those equipments in a global information system. After this analysis, architecture of the application *BW-Eye* was specified:

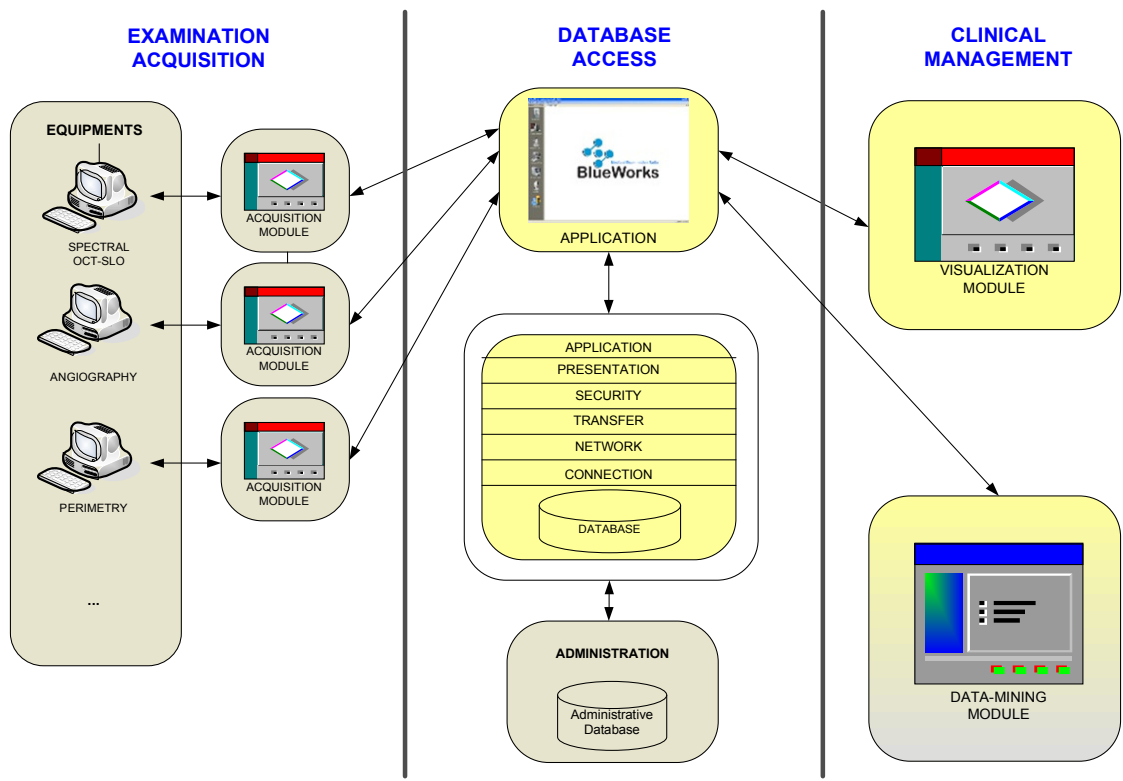

**Diagram 3 – *BW-Eye* architecture (adapted from an internal document of *Blueworks*)**

The architecture of *BW-Eye* can be divided in three fundamental components: exam acquisition modules, application database and clinical management modules.

### 1.1.2.1 Exam acquisition modules

Middleware composed by several modules that establish the interface between the examination equipments and the other components of *BW-Eye* application. Usually, each equipment has its own characteristics, thus, it was developed a specific software module to integrate the data and information produced by each examination equipment. The software modules for some of the equipments, were already developed and in the scope of a similar trainship. Liliana Ferraz has developed the software module for the retinal angiography equipment.

### 1.1.2.2 Application database

This component of the application stores the data imported from the examination equipments and other data inserted in *BW-Eye* application or generated by it as a result of some type of data processing. The different types of data are usually associated to one patient and can be accessed by users with different user profiles.

### 1.1.2.3 Clinical Management

Component of the application which allows the assessment by users to the data of patients and information (results of exams, reports and clinical data). This component implements some functionalities that allow the extraction of additional information and knowledge from the raw data and information produced by equipments, doctors and other staff. One of those functionalities is a data mining module which will be the core of the decision support system for ophthalmology. This module will be responsible by the implementation of the functionalities already indicated in this document, which will improve the traditional workflow in the consultation of the patients in a clinical ophthalmology service. Another functionality that the *BW-Eye* will implement is a software module which will allow the visualization of images, apply simple operations to the images (change the contrast of images, luminosity, zoom operations, resize images) and it will be able to do image registration: align images of the same modality, different modalities, image mosaicking, animations.

## 1.2 Objectives

The work reported in this document has as objective the implementation of an image registration algorithm integrable in the visualization module of the *BW-Eye* application. More details about the work requirements are indicated in the section 3.1 – Project Requirements.

## 1.3 Document structure and organization

This document is constituted by six main sections. In the first one, the introduction, it is described the context in which the reported work arise, the previous developed tasks and the objectives of this work. In the second one, the project management, there are identified the persons involved in this work and the projected task distribution of this work. In the third one, project analysis, there are identified the project requirements and there are reviewed some theoretical concepts related to this work subject. In fourth one, the algorithm implementation, there are described the various steps that constitute the implemented algorithm. In the results and discussion section there are presented some result obtained. The last section is the conclusions and future work section.

# 2. Project Management

## 2.1 Project team members

The project team was comprised of three Biomedical Engineering students of the Faculty of Science and Technology of University of Coimbra, project supervisors and co-workers.

Table 2- Project team members

| Name | Designation | Contact |
|------|-------------|---------|
| Liliana Ferraz | trainee student | lilianajsferraz@gmail.com |
| Ricardo Martins | trainee student | ricardo.f.a.martins@gmail.com |
| Rui Maricato | trainee student | rui.maricato@gmail.com |
| Prof.º José Basílio Simões | project coordinator | jbasilio@isa.pt |
| Drº António Travassos | Supervisor | info@ccci.pt |
| Eng. Lara Osório | Supervisor | losorio@isa.pt |
| Eng. Jorge Saraiva | Supervisor | jsaraiva@isa.pt |
| Eng. Armanda Santos | Engineering Collaborator | asantos@blueworks.pt |
| Eng. Edgar Ferreira | Engineering Collaborator | eferreira@blueworks.pt |
| Eng. Paulo Barbeiro | Engineering Collaborator | pbarbeiro@blueworks.pt |

## 2.2 Project supervising

The trainee students were integrated at *Blueworks – Medical Expert Diagnosis, Lda*, where their daily work  was followed by Engª Armanda Santos, Engº Edgar Ferreira and Engº Paulo Barbeiro. *Blueworks* is the first spin-off promoted by students of the Biomedical Engineering course of University of Coimbra. The company wants to develop innovative support decision systems for ophthalmology and other applications to help the reduction of medical subjectiveness.

Weekly, it was made a written report that described the tasks developed during that week. Those reports were sent to the elements of the project team. When the board of directors and/or the members of the project team considered opportune, the trainee students had taken part in their meeting to do small presentations about the tasks that were being developed or to participate in technical discussions.

## *2.3 Planning*

**Table 3 - Initial tasks planning**

|   | Task | Begin | End | Duration |
|---|------|-------|-----|----------|
| **1** | State of the Art retinal image registration | 14-09-2007 | 15-11-2007 | 45 days |
| **2** | Feature detection | 19-11-2007 | 08-02-2007 | 60 days |
| **3** | Features characterization | 11-02-2007 | 04-04-2008 | 40 days |
| **4** | Features matching | 07-04-2008 | 30-05-2008 | 40 days |
| **5** | Transformation model parameters estimation | 02-06-2008 | 20-06-2008 | 15 days |
| **6** | Images transformation and visualization | 23-06-2008 | 11-07-2008 | 15 days |

During the project, there were two public presentations promoted by the Biomedical Engineering course coordinators in which the work developed by the trainee students were presented.

During the first semester the project has been developed simultaneously with the attendance of more two courses: Medical Informatics and Tele-Healthcare Systems; Neuronal Computation and Diffuse Systems.

# 3. Project Analysis

## *3.1 Project Requirements*

Analyzing the studies [1] [2] [3] about the tasks and procedures developed in *Centro Cirúrgico de Coimbra* it is possible to identify some of them related with the reports of the exams. Usually in *Centro Cirúrgico de Coimbra*, after the request of a doctor for an exam of a patient, the exam is made by a technician that produces a report in a paper format. The report is given to the patient that delivers it to the doctor to be analyzed. These kind of reports are produced when the doctor requires exams such as retinal angiography, retinography (color and/or red-free), OCT (Optical Coherence Tomography) and they consist in a group of some sheets of paper with information about the patient and some images of the exam selected by the technician printed on it.

The integration of the image registration algorithm in the visualization module of the *BW-Eye* application will provide a complementary way to the traditional paper reports of analyzing the results of the exams required by the doctor. The reports of an angiography exam consist in some images selected by technician from the set of images acquired by him. Although only few of those images are selected to the report, all of them are saved in the *BW-Eye* database. During the analysis of exam results, the doctor may have some doubts and may want to visualize all the images acquired and saved during the examination. The doctor may want to visualize each image individually or the he may want to reconstruct the dynamic of the evolution of the distribution of the dye during the examination based in the images acquired. To implement this functionality it is necessary that the registration algorithm has the capacity to produce a globally aligned and ordered sequence of angiographic images.

As the reports of angiography exams, the reports of retinography (color and/or red-free) consist in some sheets of paper with images printed on them. Sometimes the region of the retina to be analyzed in detail has an area that demands the acquisition of more than one image. In these situations the images acquired individually are printed separately. When the doctor analysis these type of reports, he has to look at each one of the images and try to combine and build the *puzzle* in his mind or imagination. The images selected to be in the report were also saved in *BW-Eye* database. It will be useful to the doctor that the visualization module has the ability to combine de images acquired individually (with some percentage of overlap) and produce a unique image of that region of retina. This tool will provide a more intuitive and integrated way to analyze that region of retina. The image registration algorithm must have the ability to register sets of images of color retinography or sets of images of red-free retinography.

The registration algorithm to be implemented in the visualization module must also has the ability to register images of different modalities (color retinography, red-free retinography; angiography) in order to provide a useful tool to combine the information given by each modality and extract additional information that can not be extracted  from the analysis of each modality individually.

## 3.2 Theory Background

In this section some theoretical topics about the subjects referred in this document are introduced.

### 3.2.1 The human eye anatomy

The human eye [4] [5]] [6] is a spherical structure with a diameter about 2.5cm. The wall of human eye consists in three layers: the outer layer (*tunica fibrosa*) is composed by sclera and cornea; the middle layer (*tunica vasculosa*) is composed by choroids, ciliar body and iris; the inner layer by the retina and the retinal pigment epithelium.

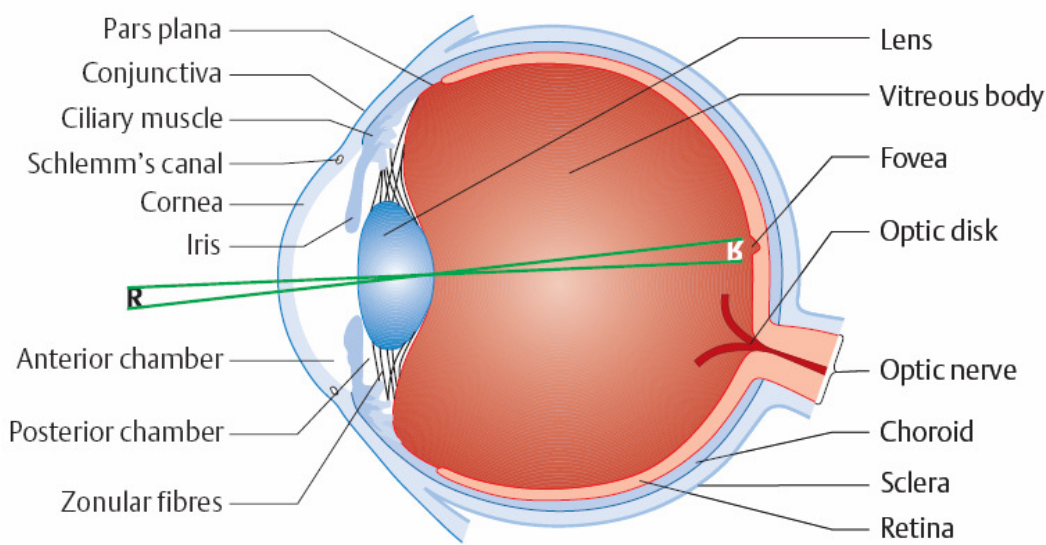

**Figure 1 – Schematic representation of the basic anatomy of the human eye [4].**

Sclera is the outer coat of the eye; usually it is white and opaque and covers 5/6 of the surface of the human eye. Sclera helps in the maintenance of the spherical shape of the human eye, it is responsible by the protection of the inner structures and it establishes the insertion point of the muscles that move the eye. Cornea is a non-vascular and transparent structure through which light is admitted into the eye.

Choroids occupies the major part of the middle layer of the eye (posterior 3/5 of the eye ball) and it contains a large amount of blood vessels which provide the blood supply (supply retinal cells with necessary oxygen and remove the waste products of the metabolism) for  the outer retinal layers. Choroid is also a pigmented tissue layer which gives the inner eye a dark color, which prevents reflections of light within the eye. Choroid is continuous with the ciliary body and with the iris.  Ciliary body extends from the end of the anterior portion of the choroids as far as the base of the iris and surrounds the iris like a ring. Lens is a biconvex and transparent structure connected by ligaments to the ciliary muscle. The lens in conjugation with the cornea helps to focus the light that enters in the eye in the retina. The shape of the lens is changed by the ciliary muscles in order to change its focal distance. Thus, objects at various distances can be focused correctly in the retina.

Iris is the visible colored part of the eye. It is a muscular contractile structure that surrounds an aperture called pupil. Pupil is a circular opening in the centre of the iris that adjusts the amount of light entering the eye. Its size is determined by the iris.

The retina is the inner layer of the eye and it is divided in a non-sensory part and an optic part: The non-sensory, the posterior part of the retina, does not have any sensory epithelium, instead of that, it has a pigmented epithelium – retinal pigmented epithelium which transports metabolic wastes from photoreceptor to the choroids. The anterior part of the retina includes the photoreceptor cells (light sensitive nerve cells) and other nine layers. The photoreceptor cells can be of two types: rods or cones and the distribution of each of these types of receptors vary across the retina. Cones are responsible for high visual acuity and color vision (photopic vision). Rods work best under conditions of reduced lighting and they neither provide well defined images nor contribute to color vision (scotopic vision). Macula lutea is a small yellow pigmented region of retina, which has in its centre the fovea, the region of greatest visual acuity. At the centre of fovea there are only cone cells. Around the fovea both rod cells and cone cells are present, with cone cells becoming fewer toward the periphery of the retina. The optic disk is a region in the retina where occurs the insertion of the optic nerve, which connects to the nerve cells and the insertion of the blood vessels that spread across the retina surface and choroids. In this location does not exist any type of photoreceptor cells, thus the optic disk location is usually called the blind spot. The retina, macula, choroids and optic disk are sometimes referred as the retinal *fundus* or *fundus*.

The vitreous cavity is a relatively large space delimited by the posterior part of the lens and the retina. The vitreous cavity is filled with a transparent material called vitreous humor, which is enclosed in the hyaloid membrane. The pressure of the vitreous humor keeps the eye ball in a spherical shape. The anterior chamber is the space between the posterior part of the cornea and the anterior part of the iris and contains a fluid called aqueous humor. The posterior chamber is a small space between the iris and the vitreous cavity.

### 3.2.2 Retinal Angiography

Retinal angiography [7] [8] is a diagnostic procedure used by ophthalmologists to photograph the retinal *fundus* and it is essentially useful to identify the blood vessels that are damaged. This technique consists in the injection of a small amount of dye in the circulatory system through a vein in the arm and then the dye travels until it reaches the vessels in the retinal *fundus*. During a period of time after the injection of the dye, pictures of the retinal *fundus* are taken in order to identify the vessels filled by the dye and to observe the dynamics of the temporal evolution of the distribution of the tracer in the retinal *fundus*. More details about the image acquisition process will be described later. The dye is eliminated by the natural metabolism of the body in about 24 hours.

There are two types of retinal angiography which depend on the type of dye used: fluorescein angiography and indocyanine green angiography. Fluorescein angiography is performed with a dye called sodium fluorescein and indocyanine green angiography with a dye called indocyanine. To carry out these exams it is necessary that the head of the patient is placed in an instrumentation in front of the *fundus* camera. This instrumentation holds the head of the patient still and places the eye in an appropriate position to be photographed. The dye is injected and a sequence of images is taken. The acquisition process varies depending on the type of retinal angiography being performed.

During the acquisition of images of fluorescein angiography, the eye is illuminated with blue light that contributes for maximum excitation of the fluorescein dye. When excited the fluorescein fluoresces in the yellow-green wavelengths. The light emitted is filtered through a filter that absorbs the blue light used to illuminate the retina and allows the yellow-green light to reach the camera (black and white film camera or a digital camera). The images acquired are grey scaled images, the first ones, taken before the dye reach the vessels of the retina, are almost completely dark images. After the dye reach the vessels of the retina, as the dye circulate in those vessels, it begins to be possible to identify the vessels in which the dye circulates. Those vessels appear in the images as white regions standing out from the darker background. The images acquired immediately after the end of the circulation of the dye in the retinal vessels are characterized by a dark background; some diffused white regions near the borders of the vessels where the dye had circulated. These diffused white regions are caused by a natural process of leakage of small amounts of dye though the boundaries of the vessels during the circulation of the dye. There are some other regions that stand out from the darker background of the images such as the optic nerve head. Those regions fluoresce due to its auto-fluorescence which occurs without the presence of any type of dye.

Several eye disorders, such as diabetic retinopathy, affect the retinal circulation and are usually studied through the fluorescein angiography. Other disorders, such as age-related macular degeneration are caused by leakage deeper choroidal blood vessels. In these cases it is necessary to do an indocyanine green angiography in order to have additional information, which is not available through fluorescein angiography.

The differences between fluorescein angiography and indocyanine green angiography are the characteristics of the dye that is used. The indocyanine green reaches its maximum excitation when it is illuminated by infrared light and fluoresces infrared light too. This characteristic allows that the infrared light emitted from the choroids vessels reaches the camera that is acquiring the images. The infrared light is not visible, thus it can not be imaged with photographic film. Instead of that, there are used high-sensitivity digital cameras in infrared spectrum.

### 3.2.3 Retinography

Retinography [9] is a technique used by ophthalmologists, which consists in the acquisition of color photographs of the retinal *fundus*. The photographs are acquired using a *fundus* camera (photographic film or digital support) with the head of the patient placed in an apparatus that holds the head of the patient and places the eye in an appropriated position to be illuminated with white light and photographed. The acquired images are color images which allow the ophthalmologists to analyze the color characteristics of the visible in the retinal fundus.

Sometimes, instead of illuminate the eye with white light, it is illuminated with green light (white light with the red component filtered). This type of retinography is called red-free retinography. The procedures are the same used in the color retinography. The images acquired are almost completely dark images with the vessels in darker colors than the retina surface (background) and the optic disk region is near white colored. The images are gray scale images. The red-free retinography provides better contrast between the vessels and the retina surface than the color retinography.

## 3.2.4 Image Registration

Image registration [10] [11] [12] is the process of transforming one or more different images of the same scene into the same coordinate system. The different images of the same scene can be acquired at different times, from different viewpoints or with different image acquisition systems, what causes that each of them will be in a different coordinate system. The objective of image registration is to determine and apply spatial transformations functions to each of the sensed images (images that are re-sampled to be registered in the reference image) in order to align each of them with the reference image (image from the set of images to be registered which is kept unchanged and defines the reference coordinate system). Each of the transformation functions maps each of the sensed images to the reference image. When the registered images are represented in the same coordinate system, the points of the different images that correspond to the same object in the scene are coincident.

Image registration is a subject with a wide range of applications [12]: cartography, industrial inspection, aerial image analysis, virtual reality and medicine. In the medical field the image registration subject is usually associated with the registration of images of CT, MRI, PET, SPECT, fMRI. However, there other applications of image registration related to medicine such as: the registration of exams with images of anatomy atlas and the registration of images of ophthalmology exams images. Usually, in ophthalmology, image registration techniques are applied in the registration of images of different modalities. Some of those applications are described in section *Project Requirements*.

It is expected that in some situations the number of images in the set of images to be registered is bigger than two. This is an important consideration, because it influences the generalization requirements of the algorithm. The principles that support the choice of the reference image and the determination of the transformations between the reference image and each of the sensed images implies the determination of the transformation between each pair of images that belong to the set of images. As it will be explained in *3.2.4.2*, the determination of the transformation between a pair of images implies the establishment of a number (variable number depending of the image transformation model considered) of correspondences of points between the two images.

The correspondences between some points of two images can be established manually. It is possible that a person looks to the two images (retinography or angiography) of one patient and he can identify homologue structures in the images that correspond to the same point in the retina (scene). Based on that correspondences, the spatial transformation of a pair of images can be determined and if necessary, the spatial transformation between the remaining possible pairs of images. However, it is possible to establish those correspondences automatically through computerized methods. The algorithm to be developed must be fully automated therefore the correspondences will be established automatically.

### 3.2.4.1 Challenges

There are several peculiarities and challenges that must be taken into consideration in order to develop an automatic registration algorithm for images of retinography (red-free and color) and retinal angiography [12] [13] [14].

The acquisition of retinography and retinal angiography images, the retina must be illuminated with the sources of illumination reported earlier. The characteristics of the techniques and the instrumentation that are employed to acquire the images may cause that some regions of the retina are illuminated differently during the acquisition of one image. It is also possible that regions of the retina are illuminated differently during the acquisition process of different images of an image modality exam session. This causes that one region of the retina may be represented by different image intensities in different images of the same modality.

In the images of healthy retinas the blood vessels (with very different diameters) are the predominant structures that standout from the relatively homogeneous background. Their density in the retinal images decreases in regions of the periphery of the retina (far from the optic disk). The existence of images of the periphery of the retina in the set of images to be registered can difficult the image registration task because usually that images have possible points of reference (retinal structures). Image sets that have low overlap percentages between images can also difficult the image registration task.

Another problem that must be taken in consideration is the presence of some diseases or lesions in the main structures of the eye that are involved in the image acquisition process. There are some diseases that may cause hemorrhages in the vascular structure of the retina, what may change the structure of the retinal blood vessels and causes the appearance of large homogeneous regions in the images that cover the retinal structures. Other ones may cause changes in the normal behavior and characteristics of the structures of the eye that are involved in the transmission of light into the retina, which may have consequences in the quality of the acquired images.

Usually the set of retinal angiography images contains some images acquired in the earlier stages of the acquisition session and some of them acquired in the later ones. In the first situation the vessels are not filled with the dye, thus the vascular structure of the fundus of the eye is not enhanced. In the second situation, the circulation of the dye already finished and the regions near the vessels borders appears as white diffuse regions (homogeneous).

Retina is a curved structure of the eye with depressions in the macula region and in the optic disk region. The image transformation between pairs of images must take this in consideration.

### 3.2.4.2 Approaches

In order to determine the transformation parameters of the retinal image registration algorithms, there were developed some different registration techniques which try to solve the image registration problem with rigor, exactness, simultaneously trying to overcome the particular challenges of the retinal image registration indicated before and trying to improve the generalization capacity of the algorithm. Although, there are several ways to classify the image registration techniques, the most used in the papers read define two classes: area-based techniques and feature-based techniques [15].

Area based techniques work with the images without attempting to extract any kind of features from the images. The methods of these techniques deal with the raw values of the pixels intensities and generally try to establish the direct correspondence between regions of the reference image and the sensed images. These regions can be rectangular (or other shapes) windows of pre-determined size. The correspondence is established based in the optimization of some objective function such as cross-correlation [16], mutual information [17]. Due to the dependence of these methods in the intensity values of the images, they are quite open to the influence of the variation of those values between images. Thus these methods are sensitive to the intensity changes introduced for instance by non-uniform illumination or pathologies. Another problem that can worsen the area-based techniques performance is the relatively large homogeneous regions in retinal images that do not have any details, structure (variable intensity values of the pixels). Large homogeneous regions are quite similar and it is quite probable that some of them are matched incorrectly with some other homogeneous region of the sensed images.

Feature-based techniques are based in the extraction of structures of the retinal images, which are designated by image features. The feature-based techniques try to replicate the manual selection and correspondence of the interest structures of the manual image registration methods. These techniques are dependent in the existence of stable and repeatable image features correspondences available between image set to be registered. It will be given a detailed explanation about the different steps involved in typical feature-based techniques because it is a more attractive approach for achieving and fulfill the objectives of this work.

The majority of the feature-based multi-image techniques consist of the following steps: image feature detection, pairwise feature matching, pairwise transformation model estimation, reference image selection, sensed images re-sampling and transformation. If there are only two images in the image set, the steps are the same, excluding the reference image selection step which is not implemented. The area-based techniques conceptually consist in the same steps of the feature-based ones, excluding the feature-detection step. Some of area-based concepts are applied in some of the steps of feature-based techniques. Usually during the development of the image registration algorithms, there is an additional step called performance evaluation, in which there are make several tests in order to evaluate the quality of the registered set of images.

### 3.2.4.2.1 Feature Detection

The first step in the feature-based registration techniques is the automatic detection of salient regions, structures or points in the images. Some approaches detect anatomical structures of the fundus of the eye that are visible in the images: blood vessels [18] [19], optic disk [20] [21] or macula [22]. Other approaches determine different types of features based in the determination of the blood vessels structure such as: blood vessels centerlines (set of pixels of one pixel width skeleton of the blood vessels structure), cross-over points and bifurcation points [12] [13] [15] [23] [24] [25]. The bifurcation points are usually called y-features and it is the type of features more used in the papers reported. There is a tendency to choose features related with the vascular structure of the retina because it is usually a stable structure and it is spread across the retina surface. Sometimes the appearance of diseases in the structures of the eye fundus can affect its stability. The position of the vessels can change due to the evolution of the disease or the vessels can become not clearly visible through the type of images already reported in this document. Some approaches such as [26] [27] try to get round this problem and develop a feature-based technique that identify and characterize by a 128 dimensional vector blob-like interest points in the highly similar regions.

### 3.2.4.2.2 Pairwise feature matching

This step consists in establishment of correspondences between the image features detected in the previous step. When the image set has more than two images, this step is repeated for all possible pairs of images. When the image set has only two images it is repeated once. In both situations for each pair of image, one of them is designated as the reference image and the other one, as the sensed image. Some approaches associate a descriptor with each of the features detected. There are some proprieties that a descriptor in a retinal image registration technique should respect which are common with other applications of feature-based image registration techniques [28]. Usually the descriptors that are used try to be invariant to the expected changes that may occur in the images to be registered. When the images are acquired under conditions slightly different from the expected ones, the descriptors must be quite similar to those that were expected if the acquisition conditions were the expected ones. Simultaneously, the characteristics of the features used in the descriptor should be specific to that feature in order to different features have different descriptors. The characteristics used in the descriptor should be independent to avoid the redundancy of the description.

The bifurcation points (y-features), cross-over points of the vessels and the vessels centerlines are the types of feature that are mostly used in these techniques. Usually the descriptors associated to these type of features are characterized by some kind of measure related with the angles between the branches of the y-feature, cross-over, the position of the centre of the y-feature, the ratio between the width of the arms of the y-feature. Typically, it is calculated a similarity measure for each possible correspondence between the features of the images of the pair. The similarity measure is based in the characteristics of the descriptor of both features and the type of similarity measure varies with the approach considered. The similarity measure can be a distance between the description of the two features in the space of the descriptors [27] or can be a more complex measure that indicates a number that suggests a more probable or improbable correspondence [12] [29] [30]. There are other approaches that do not rely on the association of a descriptor with the features. Some of those approaches apply methods inspired in the area-based registration techniques to establish the correspondences. In [24] [25] it is determined the mutual information of a pre-determined size region of the images that enclose each feature of the pair of Y-features. The pairs of features with maximum mutual information are correspondents.

### 3.2.4.2.3 Pairwise Transformation Model Estimation

For each of the possible pair of images in the set of images, after the determination of the features matching, it is estimated the spatial transformation that registrates that pair of images. The estimation of the parameters of the spatial transformation firstly consists in the selecting the type of transformation and then the estimation of its parameters. There are different types of transformations [15]. Some of them which are applied in the retinal image registration are listed in table 4.

**Table 4 – 2D Transformation models**

| type | | transformation model |
|---|---|---|
| **Linear Transformations** | Translation | $\begin{pmatrix} x' \\ y' \\ 1 \end{pmatrix} = \begin{pmatrix} 1 & 0 & t_x \\ 0 & 1 & t_y \\ 0 & 0 & 1 \end{pmatrix} \begin{pmatrix} x \\ y \\ 1 \end{pmatrix}$ |
| | Rigid | $\begin{pmatrix} x' \\ y' \\ 1 \end{pmatrix} = \begin{pmatrix} s_x r_{1,1} & s_x r_{1,2} & t_x \\ s_x r_{2,1} & s_x r_{2,2} & t_y \\ 0 & 0 & 1 \end{pmatrix} \begin{pmatrix} x \\ y \\ 1 \end{pmatrix}$ |
| | Affine | $\begin{pmatrix} x' \\ y' \\ 1 \end{pmatrix} = \begin{pmatrix} \theta_{1,1} & \theta_{1,2} & t_x \\ \theta_{2,1} & \theta_{2,2} & t_y \\ 0 & 0 & 1 \end{pmatrix} \begin{pmatrix} x \\ y \\ 1 \end{pmatrix}$ |
| **Non-Linear Transformations** | Quadratic | $\begin{pmatrix} x' \\ y' \end{pmatrix} = \begin{pmatrix} \theta_{1,1} & \theta_{1,2} & \theta_{1,3} & \theta_{1,4} & \theta_{1,5} & \theta_{1,6} \\ \theta_{1,1} & \theta_{1,2} & \theta_{1,3} & \theta_{1,4} & \theta_{1,5} & \theta_{1,6} \end{pmatrix} \begin{pmatrix} x^2 \\ xy \\ y^2 \\ x \\ y \\ 1 \end{pmatrix}$ |

$t_x$- translation parameter along $x$ axe    $t_y$- translation parameter along $y$ axe

$r_{1,1}$  $r_{1,2}$ …. $r_{2,2}$ – rotation parameters    $\theta_{1,1}$  $\theta_{1,2}$ …. $\Theta_{2,6}$ – general motion parameters

(x',y') -  pixel coordinates in the reference coordinate system

(x',y') – pixel coordinates in the sensed image

The choice by one of these transformation models should take in consideration the prior knowledge about the image acquisition conditions (instrumentation, technique), the physical characteristics of the eye and retina (anatomy, the expected extension of the retina that were photographed), the characteristics of the camera and the required and/or wanted accuracy of the registration. Among these transformation models, the quadratic transformation model, described in [12], is considered a good solution for the majority of retinal image registration challenges. In this transformation model, the retina is modeled as a rigid quadratic surface photographed by a weak perspective camera. The retina is modulated as a quadratic surface because its surface can be considered almost spherical due to the shape of the eyeball and it only has slight depressions in the region of the macula and optic disk. It is also reasonable to assume that, in common situations, the retina is rigidly attached to the back of the eye, thus, it is considered that all the points in the retina surface move uniformly. This consideration is not completely correct when the photographed retina is detached. In these cases it is suggested in [12] the application of the quadratic transformation model combined with local deformations. The application of the quadratic transformation model produces accurate results in the majority of situations as reported in [13] [31]. However, this transformation model has a high computing cost due to its complexity (estimation of 12 parameters) and requires at least the establishment of 6 features correspondences. It is usually used to register sets of images that globally correspond to large extensions of the retina because in those sets of images, the effect of the curvature of the retina has more impact between the images than in sets of images that globally correspond to small areas of the retina.

For these sets of images, the images of the set have a high percentage of overlap; there are other transformation models that can be used such as: translation, affine [15] [25], rigid body. It is considered by these models that the region of the retina that is photographed is a flat surface because it has a small extension. The translation transformation model can only handle with displacements between images. The rigid body transformation can deal with translations, rotations and scaling between images. Both of them cannot handle with images with distortions. An affine model handles translations, rotations, scaling and shearing distortions.

To solve the registration problem, it is necessary to estimate the parameters of the considered transformation, which are indicated in Table 4. Each type of transformation has a specific number of parameters. Usually it is not considered only one transformation model to register a pair of images. Many algorithms, such as [12] [15], estimate the transformations referred before using a hierarchy of transformation models and parameter estimation techniques. Simple models (translation, rigid body) are considered in the first stages of the hierarchy in order to globally align the two images. Usually the affine and quadratic transformation models are the last stages of the hierarchy. Typically the first model is estimated based in the features correspondences matched by some similarity criterion. To the other models the parameters are estimated and simultaneously the features correspondences refined.

### 3.2.4.2.4 Reference Image Selection

In a multi-image registration problem, the selection of the reference image is a very important step because it defines the coordinate system in which all the other images of the set will be registered. Usually, the reference image selection can be conceptually explained by considering each image of the set as a point and the registration of a pair of images a line which connects the two points which represent the two images involved. If the two images are not registrable, there is no line connecting these two points. Some approaches select the reference image by choosing the image (point) that connects to all the others images (points) through the minimum number of connection. Other approaches [24] associate a length to each connection. Usually this length is a measure of the error associated to the registration of the pair of images represented by that connection. This approach select the reference image by choosing the image(point) that connect to all other images (points) through the shortest path.

### 3.2.4.2.5 Sensed Images Re-sampling

After the determination of the transformations between each possible pair of images in the set of registration images, after the selection of a reference image that defines a reference coordinates system and taking into consideration the connections established by the point that represent the reference image in the previous step, it is possible to determine the transformation of any image in the reference coordinate system. After the application of the transformation to the pixels of a sensed image, it is not guaranteed that contiguous pixels in the sensed image will remain contiguous pixels after the application of the transformation. This may cause the formation of empty regions in the new registered image. The values of the pixels of those regions are obtained by interpolation of the pixels around them in the registered image.

## 3.2.5 Image Entropy

The entropy [15] [32] of a grayscale image is a statistical measure that is defined as a function of the state of probability of the image. It can be defined by the expression:

$$H^q = -\frac{1}{q} \sum_i p(s_i) \log_2 p(s_i) \qquad (1)$$

The summation is made for all possible grey level sequences of length $q$. Various orders of entropy can be defined depending on the value of $q$. The global entropy, $H^1$, is defined by the expression:

$$H^1 = -\sum_{i=0}^{255} p(s_i) \log_2 p(s_i) \qquad (2)$$

$p(s_i)$ is the probability of the occurrence of the gray level $s_i$ in the image.

The local entropy, $H^2$, is defined by the expression:

$$H^2 = -\frac{1}{2} \sum_{i=0}^{255} \sum_{j=0}^{255} p(t_{ij}) \log_2 p(t_{ij}) \qquad (3)$$

$$p(t_{ij}) = \frac{t_{ij}}{N} \qquad (4)$$

$p(t_{ij})$ is the probability of co-occurrence of the gray levels $i$ and $j$. The probability of co-occurrence is determined trough the co-occurrence matrix $\mathbf{T}=[t_{ij}]_{256x256}$. $t_{i,j}$ indicates the number of co-occurrences between the pixels $i$ and $j$ in the image. $N$ is the total number of co-occurrences in the matrix $\mathbf{T}$. $i$ and $j$ are co-ocurrent pixels if

*image(l,k)=i*    **and**  *image(l+1,k)=j*

**or**

*image(l,k)=I*    **and**    *image(l+1,k+1)=j*

### 3.2.6 Mutual information

The mutual information [15] [33] of two images $I_A$ and $I_B$ is a measure of the amount of information shared between the images. It can be defined as:

$$E_{MI}(I_A, I_B) = H(I_A) + H(I_B) - H(I_A, I_B) \qquad (5)$$

$H(I_A), H(I_B)$ are the global entropy of $I_A$ and $I_B$ respectively.

$H(I_A, I_B)$ is the joint entropy, the global entropy of the joint probably distribution of $I_A$ and $I_B$.

# 4. Algorithm Implementation

This section describes the various steps of the implemented image registration algorithm. Although in section 3.1 it is referred the objective of this work is to develop a registration algorithm with a sufficient generalization ability to register set of images with more than two images, the implementation presented is restricted to the registration of pairs of images of the modalities indicated before in this document. In the development of this implementation it was tried do develop a modular approach inspired in the fundamental steps of a typical retinal image registration algorithm, already identified in the section 3.1. This modular structure can give flexibility and versatility to the implementation and can be important in order to facilitate future modifications introduced in the approach implemented in this work.

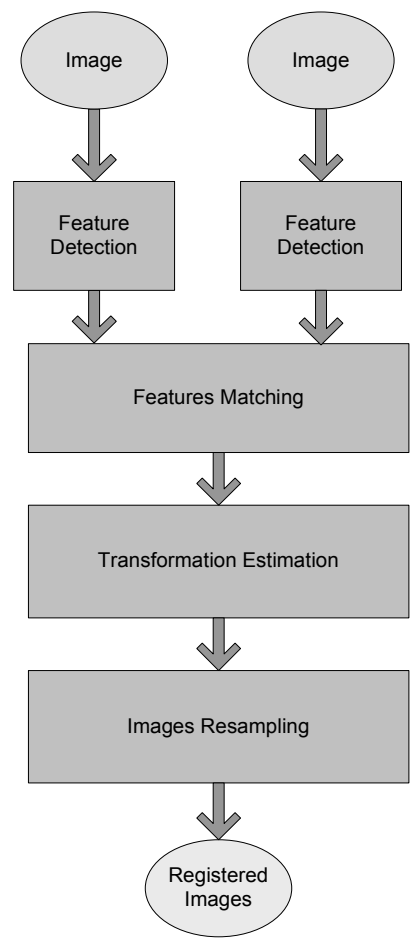

**Diagram 4 – Image pair registration algorithm overview**

This algorithm was implemented in the software *Matlab Version 7.5.0.342 (R2007b)*, running in the operating system *Windows XP*, *Home Edition*, *Version 2002*, *Service Pack 2*

## *4.1 Feature Detection*

In this step of the algorithm features of each image are determined, which will be used later in the estimation of the transformation parameters. The features used in the implemented algorithm are the bifurcation points (y-features) of the vessels of the retinal *fundus*. This step is implemented through the function `featureDetermination` described in Table 5.

Table 5 – `featureDetermination` function parameters description

| | name | description | type |
|---|---|---|---|
| **input parameters** | image | Image in which the features must be detected. | 8-bit integer bidimensional array (grayscale images) <br><br> or <br><br> 8-bit integer tridimensional array (RGB color images) |
| | modality | Variable that indicates the modality of the image given in the input parameter image. <br> 1-retinography (color) <br> 2- retinography (red-free) <br> 3- angiography | Integer |
| **output parameters** | features | Bifurcation points of the vessels. | One dimensional struct array (1 x number of features) with the fields: <br> -index <br> -branch_positions <br> -bifurcation_region <br> -index <br> -nLines <br> -nColumns |

The strategy of the implemented approach to determine the position of the bifurcation points can be described by the diagram 5.

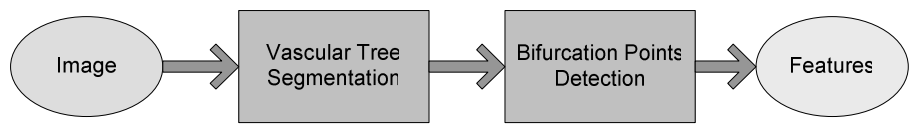

**Diagram 5 - Schematic representation of the global implementation of the function `featureDetermination`**

The vascular tree segmentation and bifurcation points determination steps are described in the following sections.

## 4.1.1 Vascular Tree Extraction

This step consists in the extraction of the vascular tree structure from the retinal images, through the utilization of a threshold dependent approach. This implementation follows the technique described in [15] to segment the vessels. Initially, it was used the original source code available at [34]. However, in the final implementation that code was edited, reorganized and in some parts, optimized. This step is implemented by the function `vasculartreeExtraction` described in Table 6.

Table 6 - `vasculartreeExtraction` function parameters description

|  | name | description | type |
|---|---|---|---|
| **input parameters** | image | Image in which the features must be detected. | 8-bit integer bidimensional array (grayscale images)<br><br>or<br><br>8-bit integer tridimensional array (RGB color images) |
|  | modality | Variable that indicates the modality of the image given in the input parameter image.<br>1-retinography (color)<br>2- retinography (red-free)<br>3- angiography | Integer |
| **output parameters** | vascular_tree | Binary image, in which it is represented the vessel structure extracted from the given image.<br>true = 0 – background<br>false = 1 – vascular tree structure | Logical bidimensional array |

| | `pre_processed_image` | Image with the vascular tree enhanced. | 8-bit integer bidimensional array (grayscale image) |
|---|---|---|---|

The approach implemented in this function works with grey level images. The red-free retinography and the retinal angiography images are already grey scale images, but the color retinography images are typically RGB color images. For this type of images, it is selected the green layer of the image to be processed by the implemented approach. The green layer is chosen because it is the layer in which the contrast between the vessels and the background is higher. The implementation of this function can be described by the diagram 6.

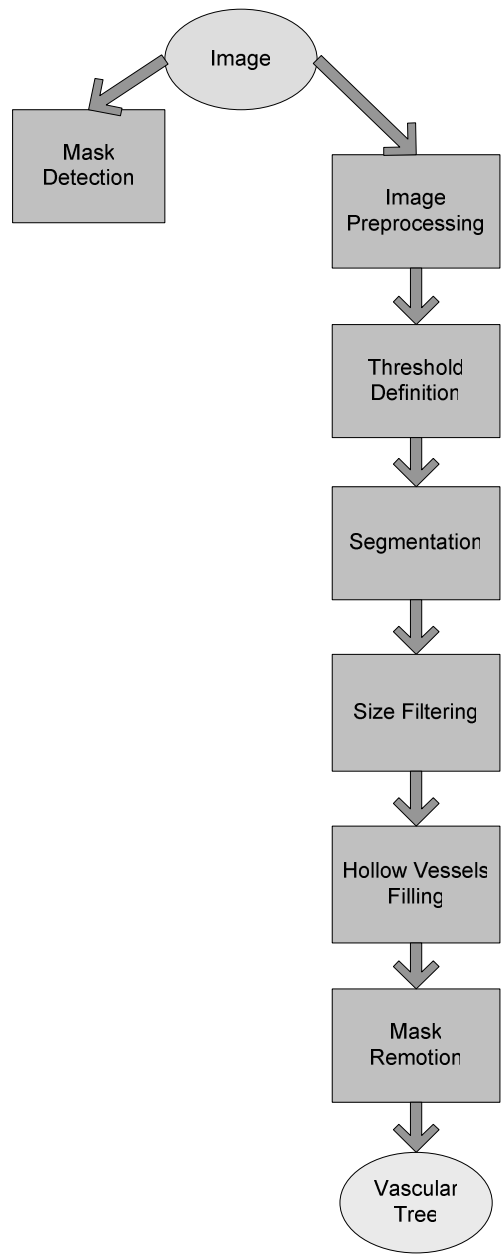

**Diagram 6 - Schematic representation of the global implementation of the function**
`vasculartreeExtraction`

### *4.1.1.1 Image Pre-Processing*

The image pre-processing step is implemented by the function `imagePreprocessing` described in table 7:

**Table 7 - `imagePreprocessing` function parameters description**

|  | name | description | type |
|---|---|---|---|
| **input parameters** | `image` | Image to be pre-processed. | 8-bit integer bidimensional array (grayscale image) |
| | `modality` | Variable that indicates the modality of the image given in the input parameter image. 1-retinography (color) 2- retinography (red-free) 3- angiography | Integer |
| **output parameters** | `pre_processed_image` | Image with the vascular tree enhanced. | 8-bit integer bidimensional array (grayscale image) |

Usually, in the type of images indicated before, the contrast between the vessels and the other retinal structures is low, thus the edges of the vessels are not ideal step. In order to enhance the contrast, it is suggested in [15] the application to the images of bidimensional matched filters. The concept of matched filters applied in the image processing field, consists in the correlation of a known template with the image. The regions of the image that have the same structure (pixel intensity profile) of the template produce higher responses in the convolution process. In [15] it is assumed that the intensity profile of the vessels along directions perpendicular to their borders can be approximated by function derivated from Gaussian functions. It is also assumed that the vessels have during a certain length (in [15] the length (L) considered is 9 pixels) a fixed orientation. The mathematical expressions implemented to define the templates that are used in this implementation are different from the ones indicated in [15], however they produce equivalent results. The fundamental template that is used to process the images with brighter vessels than the background (angiography) can be defined by the expression:

$$template(x, y) = e^{\frac{x^2}{2\sigma^2}} \qquad , y \leq \left| \frac{L}{2} \right| \qquad (6)$$

The direction of the vessel is assumed to be aligned with the *y-axis* and *L* is the number of pixels during the vessel has a fixed orientation (L=9). The value of the standard deviation (δ) of the Gaussian function is assumed to have the value 2, as indicated in [15]. The vessels in the image may have different orientations, thus there were implemented twelve different templates obtained through the rotation of the fundamental template in 0º, 15º, 30º, 45º, 60º, 75º, 90º, 105º, 120º, 135º, 150º, 165º to cover all possible orientations (with a angular resolution of 15º).

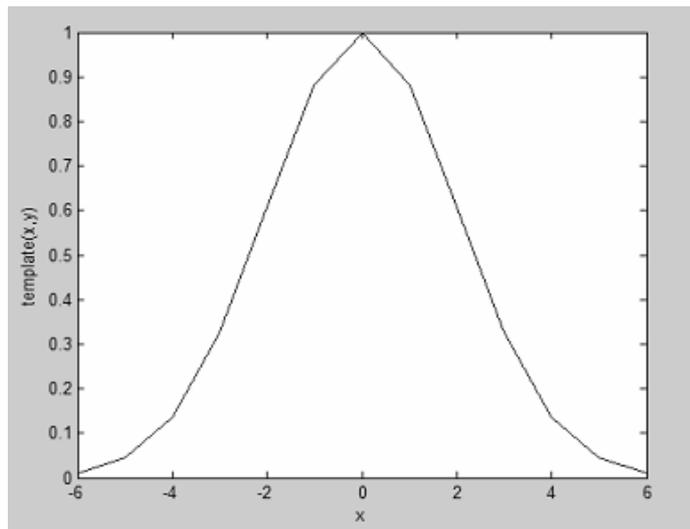

**Figure 2 – Pixel intensity profile of one line of the fundamental template defined by the expression (1). -6<x<6**

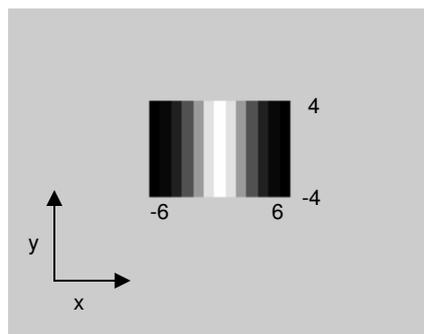

**Figure 3 – Fundamental template defined by the expression (6). -6<x<6**

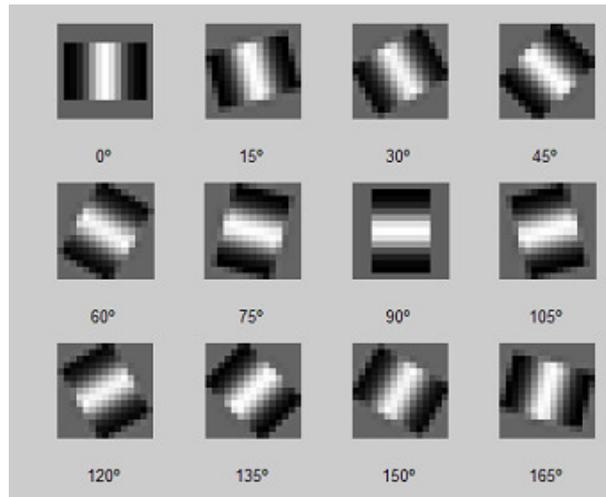

**Figure 4 – Templates used to enhance the images with bright vessels.**

The same reasoning can be applied to define a set of templates to process images with darker vessels than the background. In this situation, the fundamental template can be defined by the expression:

$$template(x, y) = 1 - e^{-\frac{x^2}{2\sigma^2}} \qquad\qquad , y \leq \left|\frac{L}{2}\right| \qquad\qquad (7)$$

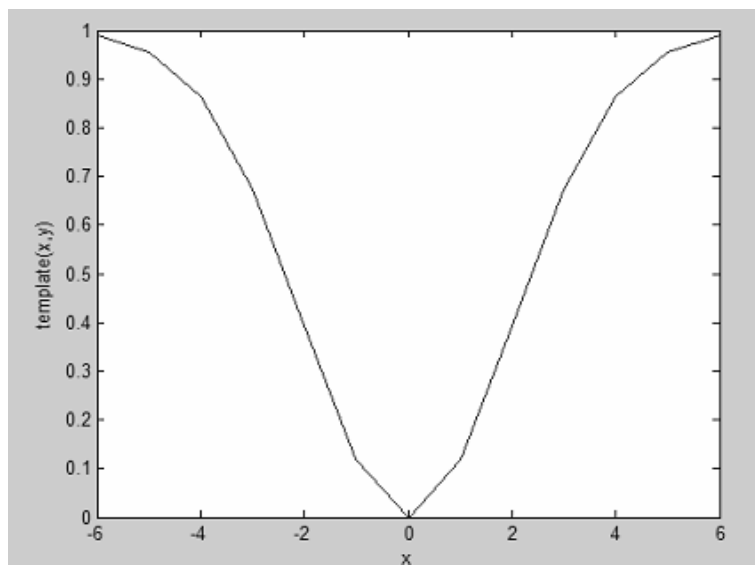

**Figure 5 – Pixel intensity profile of one line of the fundamental template defined by the expression (7).**

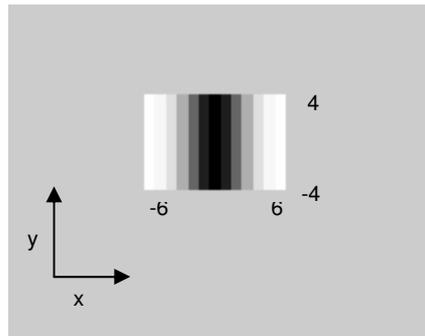

**Figure 6 – Fundamental template defined by the expression (1). -6<x<6**

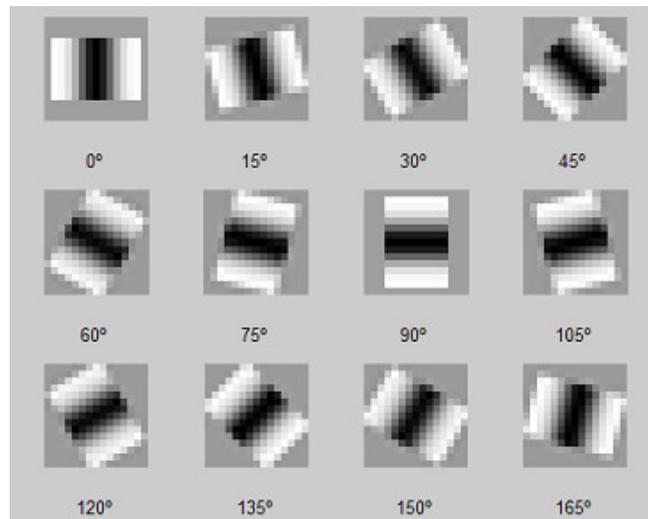

**Figure 7 – Templates used to enhance the images with dark vessels.**

The image is convolved with each one of the twelve templates of the respective set of templates. The enhanced image is obtained by selecting for each pixel the value of the maximum response between the results of all responses produced by the templates for that pixel.

### 4.1.1.2 Threshold Determination

In order to extract the vascular tree from the enhanced image, it is necessary to define a threshold (a grey level value) that establishes the classification of each pixel in background or in vessel structure. The threshold determination is implemented by the function `thresholdDetermination` (table 8)

**Table 8 - `thresholdDetermination` function parameters description**

|  | name | description | type |
|---|---|---|---|
| **input parameters** | image | Enhanced image | 8-bit integer bidimensional array (grayscale image) |
| **output parameters** | threshold | Determined threshold value. | 8-bit integer |

In [15] it is proposed the implementation of an entropy based threshold. The definition of the threshold is based in the concept of local entropy, introduced in section 3.2.3, however the co-occurrence matrix is defined in a different way. The co-occurrence matrix, **T** = [$t_{i,j}$] $_{255 \times 255}$ , of the image *Image (x, y)* is defined in [35]  by the expression:

*If*

*Image(l,k) = i   and   Image(l,k+1) = j   and   Image(l+1,k+1) = d*

*Then*

$t_{i,j} = t_{i,d} + 1$

If it is assumed that the threshold has the value *s* (0 ≤ s ≤ 255), then the co-occurrence matrix can be divided in four sections:

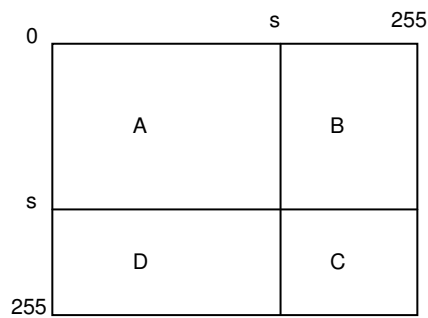

**Figure 8 – Four Sections of the co-occurrence matrix that can be considered, assuming a threshold value *s*. [35]**

The value *s*, is selected in order to maximize simultaneously the total local entropy of the sections A and C of the co-occurrence matrix of the enhanced image. The enhanced image is a bipolarized version of the original one. The pixels that correspond to the vessels structures have high values of intensity (near 255 – region C) due to the high response that they produce when are convolved with one of the templates. The regions of the enhanced image that correspond to the background have lower pixel intensity values (region A) due to the weak response in the convolution process with the templates.

The total entropy of the regions A and C can be defined as:

$$H^2(s) = H_A^2(s) + H_C^2(s) \qquad (8)$$

$$H^2(s) = -\frac{1}{2} P_A \log_2 P_A - \frac{1}{2} P_C \log_2 P_C \qquad (9)$$

$$P_A = \sum_{i=0}^{s} \sum_{j=0}^{s} p_{i,j} \qquad (10)$$

$$P_C = \sum_{i=s+1}^{255} \sum_{j=s+1}^{255} p_{i,j} \qquad (11)$$

$$p_{i,j} = \frac{t_{i,j}}{N}, \qquad (12)$$

$$N = total \quad number \quad of \quad co-ocorrences$$

The value of *s* that maximizes $H^2(s)$ is selected as the threshold value.

### 4.1.1.3 Segmentation

The segmentation step is a very simple process. The intensity value of each pixel of the enhanced image is analyzed: if the intensity value is higher than the threshold value, the pixel corresponds to a vascular structure region; if the intensity value is lower than the threshold value, the pixel belongs to the background. The result of the segmentation process is a bidimensional binary array with the pixels that correspond to the background with the value *0* (false) and the pixels that correspond to the vascular tree with the value *1* (true).

### 4.1.1.4 Size Filtering

The segmentation process can produce some small and isolated sets of pixels that were identified as vessels structures. Although that sets of pixels can really correspond to small segments of the vascular structure, the small size of these regions indicate that it is quite probable that these regions correspond to the segmentation of other structures than the retinal vessels (such as noise in the image). In the implemented algorithm the regions of the vascular structure that have a size (number of pixels) below 0.09% of the total number of pixels of the image are reclassified as background: their value is set to *0* (false). The value 0.09% was chosen taking in consideration several tests during the its implementation and considering that in [15] it is assumed a fixed value of 950 pixels and it is tested for high resolution images (950 pixels ≈ 0.0009x(1012x1024)). The identification of the different regions of the vascular structure is made through the function of the *Matlab,* `bwlabel`, using the principle of the 8-connected neighboring regions. The pixels of each identified region are labeled differently which enables the determination of the size of each region.

### 4.1.1.5 Fill Hollow Vessels

The templates defined in the image pre-processing step assume that all the vessels have the same width and thus the intensity profile of the vessels is the same. However, in some type of images the intensity profile of the larger vessels is not the expected one. In the angiography images, the larger width vessels tend to have a darker central portion in some regions of its length, which are considered background during the vessels structure segmentation step. Thus, in these situations, that segment of the vessel is segmented as two different vessels that correspond to the region of the borders of the true vessel, This will have a negative impact in the next steps of the registration algorithm, especially in the determination of the bifurcation points of the vessel. The implemented solution tries to fill the regions of the original vessel between the two borders which were segmented as two independent vessels. The implemented approach (function `fillHollowVessels`) identifies the 8-connected neighboring regions of the background and determines its size. The ones with a size below the value of 0.0115% of the total number of pixels of the image are considered vessel structures. The area value selected was chosen though some attempts during the realization of small implementation tests.

**Table 9 - `fillHollowVessels` function parameters description**

| | name | description | type |
|---|---|---|---|
| **input parameters** | vascular_tree | Binary image, in which it is represented the vessel structure extracted from the given image, with the hollow vessels not filled.<br><br>false = 0 – background<br>true = 1 – vascular tree structure | Logical bidimensional array |
| **output parameters** | vascular_tree | Binary image, in which it is represented the vessel structure extracted from the given image, with the hollow vessels filled.<br><br>false = 0 – background<br>true = 1 – vascular tree structure | Logical bidimensional array |

### *4.1.1.6 Mask Determination*

Usually, the images of retinography and angiography are fitted in by a black mask. During the extraction of the vascular tree, the inner borders of the mask are identified as vascular a vascular structure. Some of the segmented vessels contact with these inner borders that are segmented, which can be interpreted as a bifurcation or cross-over point. The points that correspond to the black mask (almost black, there are some little deviations in the pixel values intensity) are determined through the function `maskDetermination`. The pixels of each line of the image before being pre-processed are scanned from the left and from the right. Until the pixels values are higher than the pixel value intensity 15 they are considered mask.

**Table 10 - `maskDetermination` function parameters description**

| | name | description | type |
|---|---|---|---|
| **input parameters** | image | Image that will be pre-processed. | 8-bit integer bidimensional array (grayscale image) |
| **output parameters** | vascular_tree | Binary image.<br><br>true = 1 – pixels of the mask<br>false = 0 – background | Logical bidimensional array |

### 4.1.1.7 Mask Remotion

The pixels of the vascular tree that are identified simultaneously as mask are reclassified as background. The mask remotion step is implemented by the function `maskRemotion`

**Table 11 - `maskRemotion` function parameters description**

| | name | description | type |
|---|---|---|---|
| **input parameters** | vascular_tree | Binary image.<br><br>true = 1 – pixels of the vascular structure.<br>false = 0 – background | Logical bidimensional array |
| | mask | Binary image.<br><br>true = 1 – pixels of the mask<br>false = 0 – background | Logical bidimensional array |
| **output parameters** | vascular_tree | Binary image with the mask removed.<br><br>true = 1 – vascular structure<br>false = 0 – background | Logical bidimensional array |

## 4.1.2 Bifurcation Points Determination

The objective of this step is the determination of the position of the pixels that correspond to the bifurcation points of the vessels that are visible in the images. This step is implemented by the function `bifurcationpointsDetermination`.

**Table 12 - `bifurcationpointsDetermination` function parameters description**

|  | name | description | type |
|---|---|---|---|
| **input parameters** | `vascular_tree` | Binary image.<br><br>true = 1 – pixels of the vascular structure.<br>false = 0 – background | Logical bidimensional array |
| **output parameters** | `bifurcation_p oints` | Bifurcation points of the vessels. | One dimensional struct array (1 x number of features) with the fields:<br>-index<br>-branch_positions<br>-bifurcation_region<br>-index<br>-nLines<br>-nColumns |

The global implementation of this function can be described by the diagram 7.

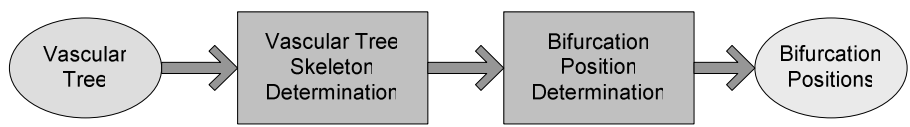

**Diagram 7 - Schematic representation of the global implementation of the function bifurcationpointsDetermination**

### 4.1.2.1 Vascular Tree Skeleton Determination

This step is implemented by the function of *Matlab*, `bwmorph`, with the parameter `operation` set to `thin`. This function reduces the vascular tree structure to a minimally connected 1 pixel width set of pixels which represent the vascular tree skeleton. It is produced a binary image in which the pixels that belong to the vascular structure tree skeleton have the value 1 (true).

### 4.1.2.2 Bifurcation Positions Determination

The pixels of the vascular tree skeleton are selected one by one. If in the 8-pixel neighborhood there are 3 or more pixels that also belong to the vascular tree skeleton, that pixel is a candidate to be a bifurcation point of the vessels. In result of this process, it is produced a binary bidimensional array with the candidate pixels labeled with the value *1* (true).

However a problem may arise. A fragment of a hypothetical vascular tree skeleton is represented in figure 9.

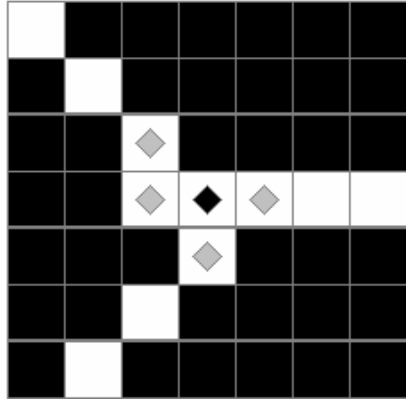

**Figure 9 – representation of a hypothetical fragment of a vascular tree skeleton.**

*White - skeleton*
*Black – background*
*Rhombus symbols – bifurcation point candidate pixels*
*Black Rhombus symbol – the selected bifurcation point candidate pixel*

Applying the criterion referred before, the pixels marked with a rhombus are possible candidates to be a bifurcation point. It is obvious that in this type of situation, only one vessel bifurcation occurs and must be considered. In order to identify all the possible situations of this type, there are detected in the binary image that has the candidate pixels, all the 8-connected neighborhood regions. Each of them forms some kind of bifurcation cluster. For each cluster it is determined its centroid (the determined coordinates are rounded). If the process is applied to the example in the figure, the chosen pixel will be the one marked with a black rhombus.

## 4.1.2.2.1 Bifurcation Positions Validation

After the procedure referred in *4.1.2.2*, it is applied to each of the candidate pixels, a procedure that tries to validate that pixel as a real bifurcation point. This procedure is implemented by the function `bifurcationpointsValidation`.

**Table 13 - `bifurcationpointsValidation` function parameters description**

| | name | description | type |
|---|---|---|---|
| **input parameters** | `image` | Enhanced image | 8-bit integer bidimensional array (gray scale image9 |
| | `Bifurcation_points_index` | Indexes of the positions of the bifurcation points candidates. | Double integer onedimensional array (line vector) |
| **output parameters** | `bifurcation_points` | Bifurcation points of the vessels. | One dimensional struct array (1 x number of features) with the fields: -index -branch_positions -bifurcation_region -index -nLines -nColumns |

The procedures consist in two types of validation. The first one selects a set of 41x4 around each bifurcation pixel candidates. If there are more than two bifurcation point candidates in that region, none of them is considered. This criterion tries to eliminate possible bifurcation pixel candidates that result from regions of the image in which the vascular tree extraction step segments noise structures that produce regions with a high density of false bifurcation pixels candidates. The second type of validation is applied to the set of bifurcation pixel candidates that result from the first one.

The second validation criterion selects a set of 41x41 pixels of the enhanced image (bifurcation region) around each bifurcation pixel candidate. Then, the profile of the values of the pixels that are sequentially positioned along a hypothetical circumference (radius=20 pixels) centered in the bifurcation point candidate pixel is analyzed. It is defined a threshold value (1,3xmean of the intensity value of the pixels selected along the circumference) that separates the intensity value profile in two main regions: the regions that whose pixels intensity values are above the threshold and those, whose pixels values intensity are below the threshold. Appling this criterion to an ideal bifurcation region, three regions with pixel values bigger than the threshold, would be identified, which correspond to the three branches of a typical bifurcation point of a vessel. For these regions, it is considered as the position of each vessel branch, the position of the pixel value with higher value on each region. However, in some situations, it will be possible that more than three regions are identified. In these situations, only the positions of the branches of the vessels that are separated at least by proximally 25º are considered (in [25], it is indicated the value 20º for the minimal separation between branches). This criterion is implemented by the function `branchpositionDetection`. If, after the application of this criterion, there are more than three possible positions for vessel branches, only three of them are selected by the function `branchSelection`. The selection is made considering a hypothetical line connecting the centre of the bifurcation region and the branch position (one of the pixels defined in the hypothetical circumference). Based in that line, it is defined a 5 pixel width rectangular region centered on them. The sum of the pixel values is analyzed and the three branches with the highest sums of the pixel intensities are selected.

For each valid bifurcation point it is saved its position index in the original full image, the position of each branch in the bifurcation region and the dimensions (number of lines and columns) of the original full image.

## 4.2 Features Matching

In this step of registration algorithm, there are established the correspondences between the feature sets of the two images. This step is implemented by the function `featuresMatching`.

**Table 14 - `featuresMatching` function parameters description**

|  | name | description | type |
|---|---|---|---|
| **input parameters** | `features_image01` | Bifurcation points of the vessels of the image 01. | One dimensional struct array (1 x number of features) with the fields: -index -branch_positions -bifurcation_region -index -nLines -nColumns |
|  | `features_image02` | Bifurcation points of the vessels of the image 02. | One dimensional struct array (1 x number of |

| | | | features) with the fields:<br>-index<br>-branch_positions<br>-bifurcation_region<br>-index<br>-nLines<br>-nColumns |
|---|---|---|---|
| | `approach` | Identification of the approach considered in the feature matching step<br>    1-approach 1<br>    2-approach 2 | Integer |
| **output parameters** | `inliers` | Position of the final matched features. | Double bidimensional array (4x(number of final matched features)) |

The implementation of this step can be divided in two parts: the first one, in which, it is established an initial set of features correspondences and the second one, in which that initial set of features is refined. The first part is implemented by two possible approaches described next.

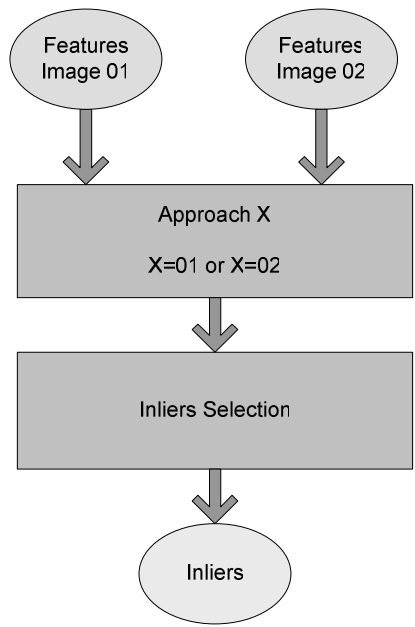

**Diagram 8 - Schematic representation of the global implementation of the function `featuresMatching`.**

### 4.2.1 Approach 1

This approach establishes an initial et of correspondences using a criterion defined in [24] and it is implemented by the function `approach01`.

**Table 15 - `approach01` function parameters description**

| | name | description | type |
|---|---|---|---|
| **input parameters** | `features_image01` | Bifurcation points of the vessels of the image 01. | One dimensional struct array (1 x number of features) with the fields: -index -branch_positions -bifurcation_region -index -nLines -nColumns |
| | `features_image02` | Bifurcation points of the vessels of the image 02. | One dimensional struct array (1 x number of features) with the fields: -index -branch_positions -bifurcation_region -index -nLines -nColumns |
| **output parameters** | `matched_features` | Position of the initial matched features. | Double bidimensional array (4x(number of features of image 01)) |

For each possible par of features between the features of image 01 and features of image 02, it is determined the mutual information of its bifurcation regions. The pair of features that have maximal mutual information is a pair of correspondent features. It was used a function called `MI2` available at [36].

### 4.2.2 Approach 2

In this approach, four invariant measures are associated with each of the features of the two images. For each feature of image 01 it is determined the feature of image 02 that is the nearest one in the invariants space. The implementation of this step is made by the function `approach02`.

**Table 16 - `approach02` function parameters description**

| | name | description | type |
|---|---|---|---|
| **input parameters** | `features_image01` | Bifurcation points of the vessels of the image 01. | One dimensional struct array (1 x number of features) with the fields: -index -branch_positions -bifurcation_region -index -nLines -nColumns |
| | `features_image02` | Bifurcation points of the vessels of the image 02. | One dimensional struct array (1 x number of features) with the fields: -index -branch_positions -bifurcation_region -index -nLines -nColumns |
| **output parameters** | `matched_features` | Position of the initial matched features. | Double bidimensional array (4x(number of features of image 01)) |

### 4.2.2.1 Invariants determination

The invariants selected to characterize each of the features are related with the angle between the branches of the bifurcations and its widths. This approach assumes that each of the bifurcation regions have a traditional structured represented in figure 10.

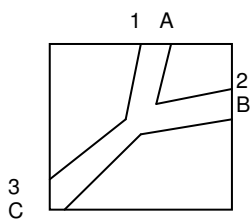

**Figure 10 – Schematic Representation of the ideal structure of a bifurcation region.**

*1,2,3 – identification of each vessel branch*

*A, B, C – width of each vessel branch*

The angles between each consecutive pair of vessel branches of the considered bifurcation point are defined by the following expressions:

$\Theta_{1,2}$ – angle between branch 1 and branch 2

$\Theta_{2,3}$ – angle between branch 2 and branch 3

$\Theta_{1,3}$ – angle between branch 1 and branch 3

The smallest angle between $\Theta_{1,2}$, $\Theta_{2,2}$, $\Theta_{1,3}$, is determined. After the identification of that angle, the widths of the two branches that define that angle are analyzed. If the widths of the branches are equal or the branch with bigger width is positioned clock-wisely to the other branch, the next angle between branches to be considered is the one positioned clock-wisely to the first one. If the branch with the bigger width is positioned anti clock-wisely to the other, then the next angle to be considered is the one positioned anti clock-wisely to the first one. The first angle indicated defines the first invariant parameter; the second angle defines the second one.

The other two invariants are obtained from the ratios of the width of the branches that define the angle indicated before. The third invariant parameter is defined by the ratio of the widths between the branch of bigger width and the width of the other branch that defines the angle of invariant parameter 1. The fourth invariant parameter is defined by the ratio of the widths of the branches that define the angle of the second invariant parameter. If the angle is clock-wisely to the angle of first invariant the ratio is between the width of the branch more clock-wisely positioned and the other. If the angle selected in the second invariant parameter is anti-clock-wisely relatively to the angle of invariant parameter 1, the ratio is defined between the branch positioned more anti-clock-wisely and the other.

In the example of the bifurcation point of figure 10, the angle $\Theta_{1,2}$ is the smallest one.

If B ≥ A

Invariant parameter 1 = $\Theta_{1,2}$

Invariant parameter 2 = $\Theta_{2,3}$

Invariant parameter 3 = B/A

Invariant parameter 4 = C/B

If B < A

Invariant parameter 1 = $\Theta_{1,2}$

Invariant parameter 2 = $\Theta_{1,3}$

Invariant parameter 3 = A/B

Invariant parameter 4 = C/A

The determination of the invariant is implemented by the function `invariantsDetermination`.

**Table 17 - `invariantsDetermination` function parameters description**

|  | name | description | type |
|---|---|---|---|
| **input parameters** | features | Bifurcation points of the vessels of one image. | One dimensional struct array (1 x number of features) with the fields: -index -branch_positions -bifurcation_region -index -nLines -nColumns |
| **output parameters** | bifurcation_invariants | Invariant parameters of the features. | One dimensional struct array (1 x number of features) with the fields: -parameter1 -parameter2 -parameter3 -parameter4 |

During the implementation of the function `invariantsDetermination` there are determined the angles and width of each branch.

### 4.2.2.1.1 Slop class determination
This function is used to determine the angle of each of the branches of a vessel bifurcation point and also return a label which identifies each of these angles with a class. This step is implemented by the function `slopeclassDetermination`.

**Table 18 - `slopeclassDetermination` function parameters description**

| | name | description | type |
|---|---|---|---|
| **input parameters** | branch_positions | positions of the three branches of a vessel bifurcation point in the bifurcation region | Double unidimenional array (1x3 line vector) |
| | dimension | dimension of the bifurcation region a vessel bifurcation point | Integer |
| **output parameters** | branchs_slop_class | slop class of the three branches of a vessel bifurcation point in the bifurcation region | Double unidimenional array (1x3 line vector) |
| | branchs_angle | angles of the three branches of a vessel bifurcation point in the bifurcation region | Double unidimenional array (1x3 line vector) |

For each branch position, it is determined the angle (by the function `equation`) that a hypothetical line that connects the bifurcation point and the branch position makes with the horizontal.

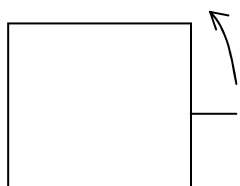

**Figure 11 – Representation of how the angle of each branch is measured**

The angle of each branch of a bifurcation point will be used in the invariant parameters determination step in order to determine the angle between each pair of branches.

This function classifies the angle of each vessel branch in one of eight possible categories defined in table 19.

**Table 19 - Categories of classification of the vessel branch angle**

| class | angle |
|---|---|
| 1 | 337.5º<angle≤360º and 0º≤angle≤22.5º |
| 2 | 22.5º<angle≤67.5º |
| 3 | 67.5º<angle≤112.5º |
| 4 | 112.5º<angle≤157.5º |
| 5 | 157.5º<angle≤202.5º |
| 6 | 202.5º<angle≤247.5º |
| 7 | 247.5º<angle≤292.5º |
| 8 | 292.5º<angle≤337.5º |

#### 4.2.2.1.2 Branches width determination

The determination of the width of the tree vessel branches of each bifurcation point is implemented in the function `brancheswidthDetermination`.

**Table 20 - `approach02` function parameters description**

| | name | description | type |
|---|---|---|---|
| **input parameters** | `Bifurcation_region` | positions of the three branches of a vessel bifurcation point in the bifurcation region | Double unidimensional array (1x3 line vector) |
| | `Branch_positons` | positions of the three branches of a vessel bifurcation point in the bifurcation region | Double unidimensional array (1x3 line vector) |
| | `Branchs_slop_class` | slop class of the three branches of a vessel bifurcation point in the bifurcation region | Double unidimensional array (1x3 line vector) |
| | `dimension` | dimension of the bifurcation region a vessel bifurcation point | Integer |
| **output parameters** | `branches_width` | Width of each of the branches of a bifurcation point | Double unidimensional array (1x3 line vector) |

For each of the branches of a bifurcation point, it is determined in the bifurcation region pixel array, the pixel positions that correspond to an imaginary line that connects the bifurcation point position to the branch position (figure 12). The positions of the pixels of that line that are near the bifurcation point are not considered.

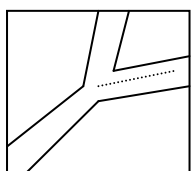

**Figure 12 – representation of the imaginary line that connects the bifurcation point to the position of one of the branches.**

For each of the pixels along that line, there are determined the pixels that correspond to a 10 pixels length line perpendicular and centered on them. The intensity values of that 10 pixels length line represent the intensity values profile of a cross section of that region of the branch. Then, it is calculated the first derivate of the intensity value profile and the positions where the maximum and the minimum values of the derivate occur correspond to the positions of the borders of the vessels: The difference between the positions of the maximum and the minimum correspond to the vessel width for that cross section of the vessel branch. The final width of the branch is the rounded mean of the determined widths for all cross sections of that branch. For branches that have a slop-class equal to 2,4,6,8, that result is multiplied by $\sqrt{2}$ .

### 4.2.3 Inliers selection

The set of matched pairs of features that results from approach 1 step or approach 2 step may have some wrong matches. In order to exclude the wrong matched pairs, it is proposed in [26] the selection of the correct matched pairs by estimating a homography [37] transformation with the RANSAC [38] algorithm. It is referred that, although the retina is not a flat surface, it is sufficient the estimation of the homography transformation parameters with a RANSAC algorithm to make a selection of the correct matched pairs of features. The algorithm RANSAC will iteratively estimate the parameters of the homography transformation model using the initial set of correspondences. As the iteration occurs, the pairs of features that cannot be fitted by the model are discarded.

Table 21 - `inliersSelection` function parameters description

|  |  | name | description | Type |
|---|---|---|---|---|
| **input parameters** | `matched_features` | | Position of the initial matched features. | Double bidimensional array (4x(number of features of image 01)) |
| **output parameters** | `inliers` | | Position of the final matched features. | Double bidimensional array (4x(number of final matched features)) |

At the end of this process, it is possible to identify which are the most probable to be correct features correspondences from the initial set of features. This step of the features matching step is implemented by the function `inliersSelection`, in which was used the code available in [39] and the presented examples.

## *4.3 Transformation estimation*

In this step of the implemented algorithm the parameters of the considered transformation model are estimated. This step is implemented by the function `transformationEstimation` and uses the pre implemented function of *Matlab* `cp2tform`.

**Table 22 - `transformationEstimation` function parameters description**

|  | name | description | type |
|---|---|---|---|
| **input parameters** | inliers | Position of the final matched features. | Double bidimensional array (4x(number of final matched features)) |
| **output parameters** | tform | Estimation of the transformation model parameters | Struct array with de structure described in the Matlab documentation |

If number of final matched features in the inliers is less than three, the registration of this pair of images is not effectuated. If it is greater or equal than three and lower than six, it is considered an affine transformation model. If it is greater or equal than six it is considered a quadratic transformation model (polynomial degree 2).

## *4.4 Images re-sampling*

In this step of the implemented algorithm is applied the determined transformation model to the *image 02*. This step is implemented by the function `imagesResampling` in which is used the *Matlab* function `imtransform`.

**Table 23 - `imagesResampling` function parameters description**

|  | name | description | type |
|---|---|---|---|
| **input parameters** | tform | Estimation of the transformation model parameters | Struct array with de structure described in the Matlab documentation |
|  | interpolation | Specifies the form of interpolation to use. 1- nearest neighbour 2- bilinear 3- bicubic | integer |
| **output parameters** | registered_images | Estimation of the transformation model parameters | 8-bit integer bidimensional array (grayscale image) or |

| | | | 8-bit integer tridimensional array (RGB image) |
|---|---|---|---|
| | | | |

## *4.5 Auxiliary functions*

This section of the document describes the functionality of some functions that were implemented and that were used to implement some auxiliary tasks in other functions.

### 4.5.1 indexesLine

This function determines the absolute indexes of the pixels of a bidimensional array that define an imaginary line that have starts and ends in the defined pixel indexes.

Table 24 - `indexesLine` function parameters description

| | name | description | type |
|---|---|---|---|
| **input parameters** | `pos_point1` | absolute position of the index of one of the two ends of the line | Integer |
| | `pos_point2` | absolute position of the index of one of the two ends of the line | Integer |
| | `nColumns` | number of columns of the array in which the indexes of the line are determined. | Integer |
| | `nLines` | number of columns of the array in which the indexes of the line are determined. | Integer |
| **output parameters** | `indexes_line` | Absolute position of the indexes that correspond to the line. | Double onedimensional array (line vector) |

### 4.5.2 equation

This function determines the slop (m) and the correspondent angle straight line of equation y=mx+b.

**Table 25 - `equation` function parameters description**

| | name | description | type |
|---|---|---|---|
| **input parameters** | `column_point1` | column index of the pixel of one end of the line | Integer |
| | `line_point1` | line index of the pixel of one end of the line | Integer |
| | `column_point2` | column index of the pixel of one end of the line | Integer |
| | `line_point2` | line index of the pixel of one end of the line. | Integer |
| **output parameters** | `m` | Slope of the line | Integer |
| | `angle` | angle that corresponds to the slop m | Integer |

### 4.5.3 indexesCircumference

This function determines the absolute indexes of the pixels of a quadratic bidimensional array (odd dimension) that define an imaginary circumference of diameter equal to the dimension of the bidimensional array.

**Table 26 - `indexesCircumference` function parameters description**

| | name | description | type |
|---|---|---|---|
| **input parameters** | `dimension` | Dimension (odd number) of the quadratic bidimensional array. | Integer |
| **output parameters** | `indexes_circumference` | Absolute position of the indexes of the pixels that define the circumference. | Double unidimensional array (line vector) |

### 4.5.4 indexesArea

This function defines the indexes of a 5 pixel width rectangular region centered in an imaginary line connecting the points given.

**Table 27 - `indexesArea` function parameters description**

| | name | description | type |
|---|---|---|---|
| **input parameters** | `pos_point1` | absolute position of the index of one of the two ends of the line | Integer |
| | `pos_point1` | absolute position of the index of one of the two ends of the line | Integer |
| | `nColumns` | number of columns of the array in which the indexes of the rectangular area are determined. | Integer |
| | `nLines` | number of lines of the array in which the indexes of the rectangular area are determined. | Integer |
| **output parameters** | `indexes_area` | Index of the pixels that correspond to the determined rectangular area. | Double unidimensional array (line vector) |

### 4.5.5 conversionindexCL

This function converts the position of an array element from an absolute format to a (line, column) format.

**Table 28 - `conversionindexCL` function parameters description**

| | name | description | type |
|---|---|---|---|
| **input parameters** | position | Absolute position | Integer |
| | nLines | Number of lines of the array | Integer |
| **output parameters** | column | Position column index | Integer |
| | line | Position line index | Integer |

### 4.5.6 conversionindexPos

This function converts the position of an array element from a line, column format to an absolute format.

**Table 29 - `conversionindexPos` function parameters description**

| | name | description | type |
|---|---|---|---|
| **input parameters** | column | Index column position | Integer |
| | line | Index line position | Integer |
| | nLines | Position column index | Integer |
| **output parameters** | position | Index absolute position | Integer |

# 5. Results and discussion

In this section of the document, there are presented some results of the application of the various steps of the algorithm described in the algorithm implementation. The tests that are made are not a traditional exhaustive set of tests, in order to evaluate the global algorithm performance (speed, generalization capacity, image registration error quantification). The tests that were made want to demonstrate the functionalities and the importance of the implemented steps in the algorithm. The tests were made in images of retinal angiography and red-free retinography acquired in *Centro Cirúrgico de Coimbra* using the fundus camera *Topcon TRC-52X*.

## *5.1 Feature detection*

In this section there are presented some results of the steps of the algorithm related to the detection of the features.

### 5.1.1 Vascular tree segmentation

The first step involved in the vascular tree extraction is the pre-processing of the image which produces a grey scale image with the vascular structures enhanced. Due to the high dimensions of the images, the results of the application of this step of the algorithm to retinal angiography and red-free images are presented in the appendices figures 1, 2, 3 and 4. The vascular structure visible in both types of images stands out from the darker background after the image pre-processing. In the enhanced image of the red-free retinography, some background structures near the right down region of the optic disk were also enhanced which can cause the segmentation of non-vascular tree structures.

After the determination of the threshold value the segmentation procedure is applied. The obtained results for the images pre-processed before are visible in the appendices figures 5 and 6. It is clearly visible in both obtained images the presence of small isolated regions of pixels identified as vascular structures. However most of them do not correspond to vascular structures in the original images. The elimination of these regions is made through the application of the size filtering. The images that result from the application of this algorithm step are also in the appendices figures 7 and 8.

The filling hollow vessels step has no visible impact in the images presented in the appendices. There are only small hollow regions that are filled. There is a hollow vessel visible in the appendices figure 8, however that hollow vessel is not filled because the hollow region has a size above the threshold defined. The final step of the vascular tree extraction is the remotion of the pixels identified simultaneously as mask of the image and vascular structure. The effect of this last step is visible in appendices figures 7 and 8.

## 5.1.2 Bifurcation points detection

The first step in order to determine the bifurcation points of the segmented vascular structure is to determine the skeleton of the vascular structure. The application of this step to the vascular tree extracted from retinal angiography images and red-free retinography images is equivalent. In figure 13 is represented a small region of a pre-processed angiography image. The pixels in blue, red and green correspond to the skeleton pixels. This colored representation is not implemented in the algorithm; it was produced only for visualization purposes for this document. The bifurcation clusters pixels are green and red colored and the centroid of each bifurcation cluster is red colored.

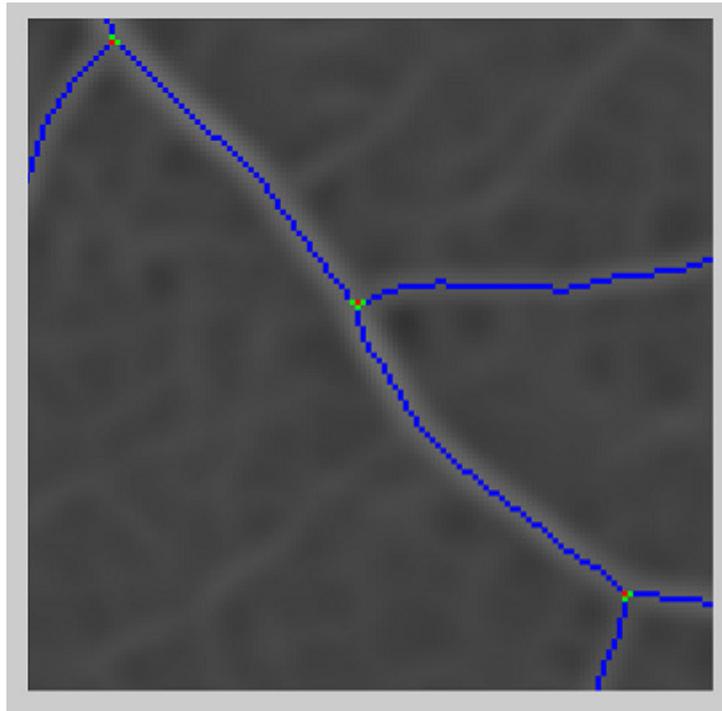

**Figure 13 - Representation of a small region of a pre-processed angiography image, which has overlapped the determined skeleton structure (blue+green+red) the bifurcation cluster pixels (green+red) and the centroids of each bifurcation cluster (red)**

The centroids of the bifurcation clusters constitute a set of possible bifurcation points. The validation of each of the centroids as a real bifurcation point is made in two steps. The application of the first one eliminates bifurcation points candidates that are in regions with a high density of bifurcation point candidates. In figure 14 it is visible a region is presented a region with those characteristics (the right down region of the optic disk of the red-free image referred before). After the application of that validation criterion the number of the bifurcation points candidates is largely reduced (figure 15). However this criterion can eliminate correct bifurcation points candidates that can exist in those regions.

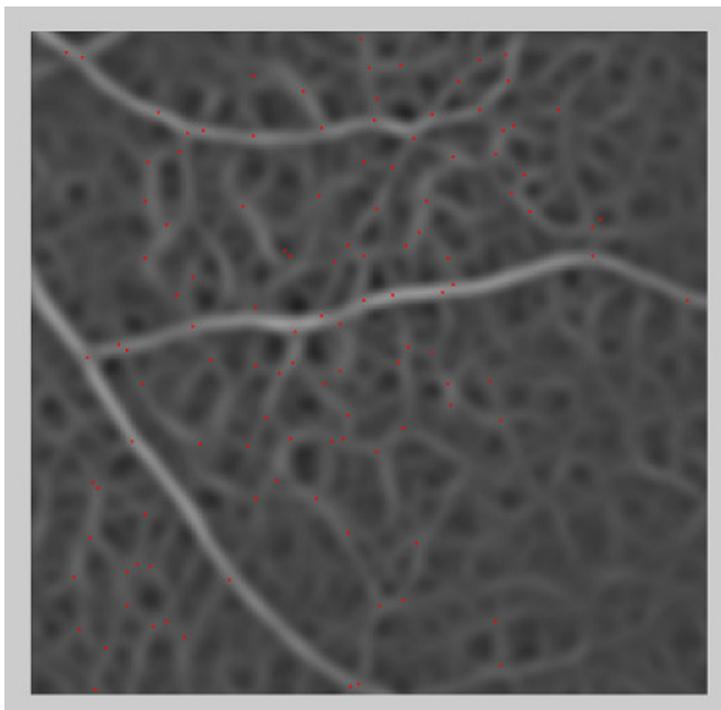

**Figure 14 - Region of a pre-processed red-free retinography image with a high density of bifurcation points candidates**

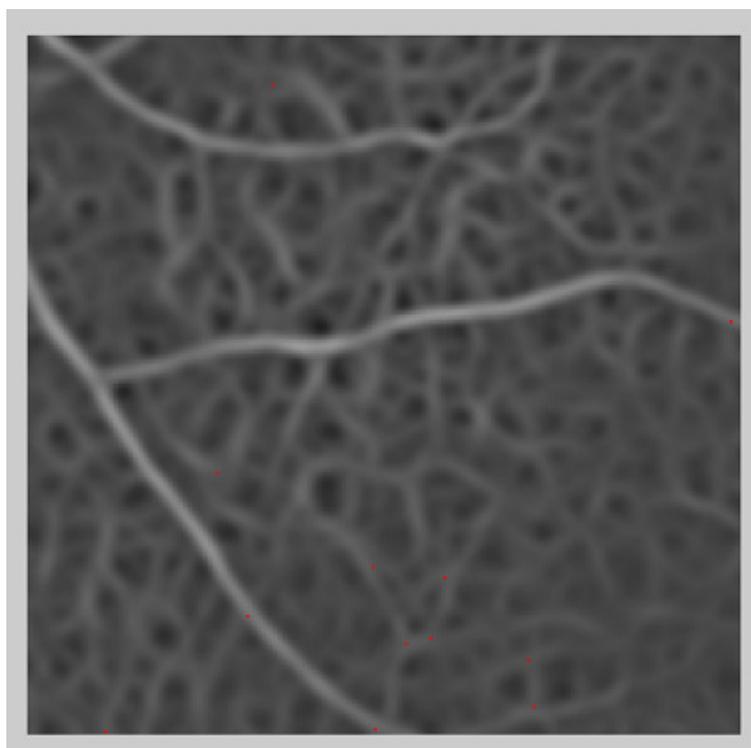

**Figure 15 - Same region of figure 15 but the bifurcation points candidates have been filtered using the first validation criterion.**

The second validation criterion tries to select only the bifurcation points candidates that have a bifurcation region with a structure similar to a bifurcation region of an ideal bifurcation point. The application of this criterion to the bifurcation point candidates of a red-free retinography image produces several typical results that can be extended to the other image modalities.

Some of the bifurcation point candidates are positioned in regions without any structure, almost homogeneous regions.

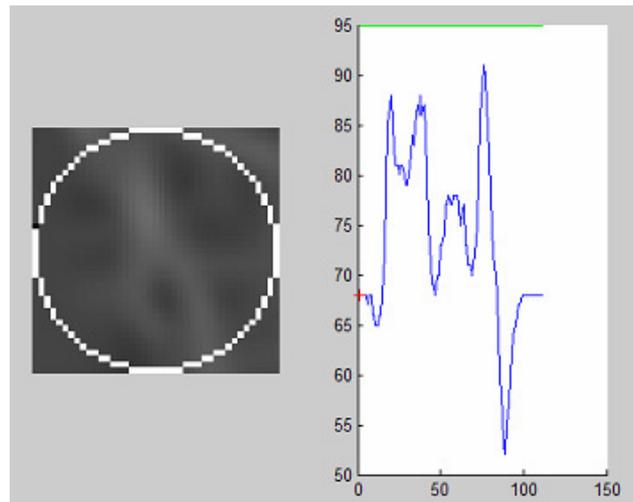

**Figure 16 - Analysis of the circumference pixels intensity value profile (blue line) around a bifurcation pixel candidate. The threshold value (green line) is not achieved**

Usually this type of bifurcation point candidate (figure 16) is not validated by this criterion, because it is not possible to identify three regions in the intensity values profiles which have an intensity value higher than the threshold. It is important to eliminate this type of bifurcation point candidates because as its bifurcation region is quite homogeneous, it is very similar to other homogeneous bifurcation regions, what can introduce errors in the features matching step.

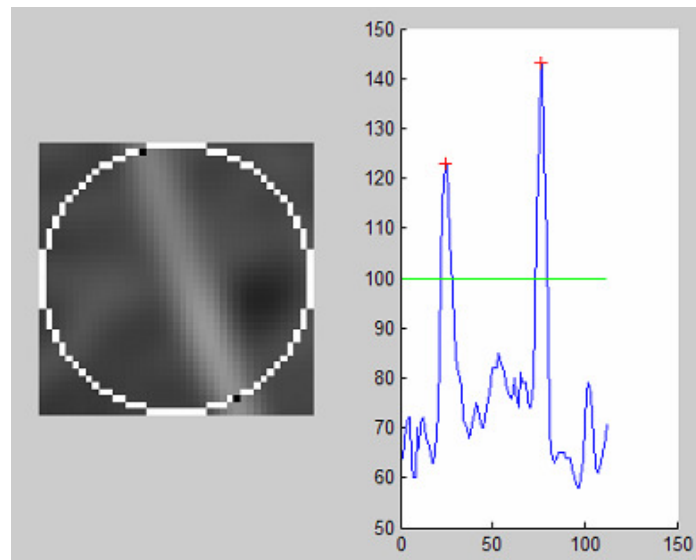

**Figure 17 - Analysis of the circumference pixels intensity value profile (blue line) around a bifurcation pixel candidate. The threshold value (green line) is achieved twice, in the regions of the interception of the vessel with the circumference.**

Other type of situations that typically may occur is the detection of bifurcation points candidate in a region that has a segment of a vessel but that segment is not a vessel bifurcation (figure 17). Usually, this type of bifurcation points candidates are eliminated because in the analysis of the circumference pixel intensity values profile the defined threshold is only achieved in two regions.

The presence of real bifurcation point (figure 18) is detected by the existence of three regions in the intensity value profile of the pixels of the circumference that have values higher than the threshold.

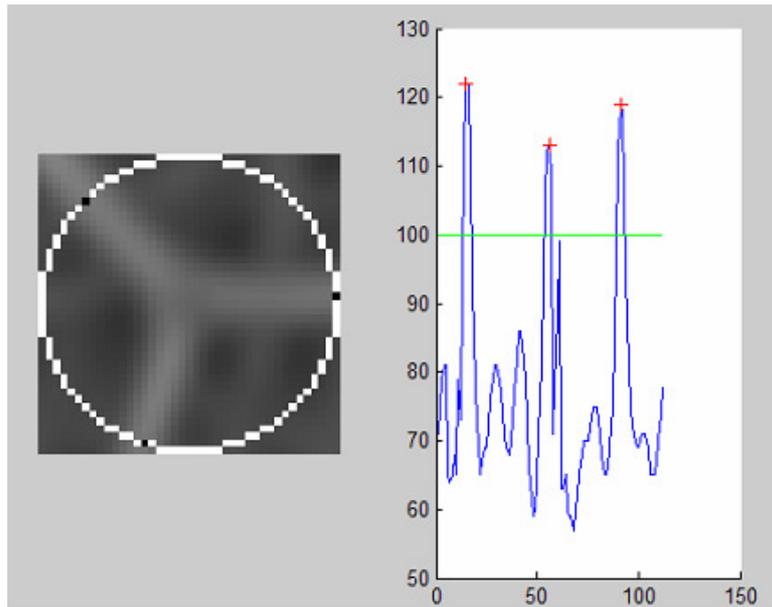

**Figure 18 - Analysis of the circumference pixels intensity value profile (blue line) around a bifurcation pixel candidate. The threshold value (green line) is achieved three times, in the regions of the interception of the branches with the circumference.**

Sometimes the threshold value is achieved more than three times (figure 19) in the circumference pixels intensity value profile. Usually, these bifurcation points candidates are validated and three of these branches are selected.

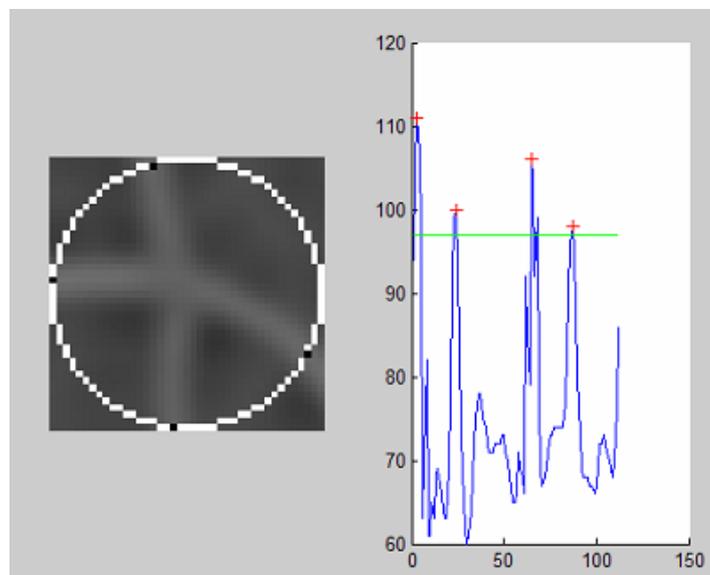

**Figure 19 - Analysis of the circumference pixels intensity value profile (blue line) around a bifurcation pixel candidate. The threshold value (green line) is achieved four times, in the regions of the interception of the branches with the circumference.**

Some real bifurcation points are not validated (figure 20) because one or more of their branches are not detected in the circumference pixels intensity values profile. This is caused by some of these branches that have a very small width or the contrast of these branches and the background in the circumference pixels positions is low.

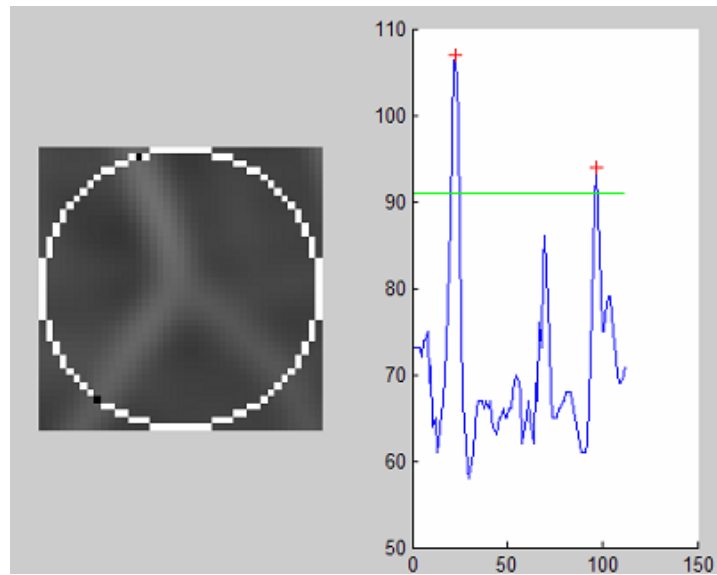

**Figure 20 - Analysis of the circumference pixels intensity value profile (blue line) around a bifurcation pixel candidate. The threshold value (green line) is achieved only two times, in the regions of the interception of the branches with the circumference. This real bifurcation point is not validated.**

For each of the images of red-free angiography and retinal angiography already indicated the feature detection step identifies 14 and 50 valid bifurcation points, respectively, that are represented in the appendices figures 9 and 10. The major parts of the bifurcation points identified in the red-free image clearly correspond to real bifurcation points. Some of the validated bifurcation points in the angiography image are not bifurcation regions.

## 5.2 Features matching

In this section, the part of the approach 2 in which the features invariants are determined will be analyzed. The remaining parts of this algorithm step will be implicitly tested during the tests of the complete algorithm. There were selected some typical situations that occurred when the determination of the invariants was tried in the set of validated bifurcation points of the angiography image already used before. The table 30 summarizes the obtained results. The white pixels that delimitate the border of the vessels branches indicate the considered vessel borders positions in the determination of the width of each vessel.

**Table 30– Invariants determination for some bifurcation points**

| Bifurcation regions | Vessels width | branches angles | invariants |
|---|---|---|---|
| 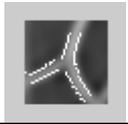 | 5<br>5<br>4 | 69.8º<br>302.9º<br>206.6º | 1.25<br>1<br>96<br>126.9 |
| 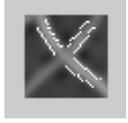 | 5<br>5<br>4 | 122.9º<br>49.7º<br>319.1º | 1.25<br>1<br>73.9º<br>163.8º |
| 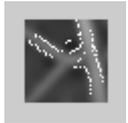 | 5<br>8<br>8 | 143.1º<br>59.5º<br>323.1º | 1.6<br>1<br>83.6º<br>96.4º |
| 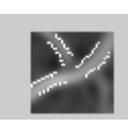 | 5<br>7<br>6 | 116.5º<br>32.9º<br>206.5º | 1.4<br>0.86<br>83.7º<br>96.4º |
| 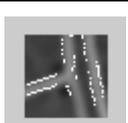 | 9<br>6<br>5 | 53.1º<br>327.1º<br>200.2º | 1.5<br>0.56<br>86º<br>147.1º |
| 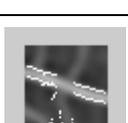 | 5<br>5<br>8 | 159.8º<br>342.5º<br>237.1º | 1.6<br>0.63<br>77.3º<br>105.4º |

The approach 2 was implemented assuming that the structure of the bifurcation region of each bifurcation point is near the ideal: the bifurcation vessel is modeled as a quite well define structure. However, as seen in the previous analyses, some of the validated bifurcation points are in bifurcation regions that sometime differ from the ideal one. This causes that the correct determination of the invariants only occurs in regions that have a structure near the ideal one. Bifurcation points that have a structure such as the third, fifth and sixth bifurcation points in the table have visible problems in the delineation of the branches borders and consequently in the determination of the branches widths and feature invariants. These errors that occur in this level of the algorithm will have a negative impact in the establishment of features correspondences.

### *5.3 Full algorithm tests*

The simplest test that can be implemented to analyze if the implementation code of the algorithm have no errors, is to test the registration algorithm with a pair of equal images. These tests were made using the angiography and red-free retinography images already referred in this document.

In the tests made using the pair of equal angiography images there were detected 50 valid bifurcation points in each of them and there were established 50 correct features matches by the inliers selection step, both when was used the approach 01 or the approach 02. The image pair was correctly registered. The same situation occurred when the tests were made using a pair of equal red-free retinography images. In each of the images there detected 14 valid bifurcation points and there were established 14 correct features matches in the in inliers selection step, both when was used the approach 01 or the approach 02. The image pair was correctly registered.

It was also tried to register pair of different images. In the first attempt, there were used two angiography images acquired in the same exam session. These two images have a high percentage of overlap; the point of view of the acquisition is near the same. In the feature detection step, there were correctly validated 49 bifurcation points for one of the images and 46 to the other. When was tried the approach01 to register this pair of images, during the inliers selection the algorithm was aborted due to the high number of initial matched features that involve the same feature of one of the images. This situation impossibly the estimation of the inliers by the RANSAC algorithm. Using the approach 02 there were not detected any inlier witch also impossibly the registration of this pair of images. There were made more tests both with more pairs of angiography images of that acquisition session (with high percentage of overlap) and with a pair of an image of red-free retinography and an angiography image. In both situations the tendency expressed in the previous results prevailed due to the same reason. In none of the tests made to the full algorithm, the pair of different images was successfully registered. However more different tests must be done.

Due to the occurrence of errors in the inliers selection step, there was made a simple test that consisted in the establishment of the manual correspondences between five bifurcation points of the two registrable images (red-free retinography and retinal angiography) and there were added more five wrong bifurcation points correspondences. The set of the five correct bifurcation points matches was determined by the RANSAC algorithm and the pair of image was successfully registered using an affine transformation model. The images original and registered images can be seen the appendices figure 11, 12 and 13.

# 6. Conclusions and future work

## *6.1 Conclusions*

Although the main global objective was not achieved there were made some analysis that can be useful in the future development and improvements of the algorithm. The implemented algorithm, to identify bifurcations points candidates, had produced good results, although in some regions there are detected a high number of false candidates due to the segmentation of structures that are not vascular structures. The segmentation method can be improved, however, is quite probable that always some structures will be wrong segmented. Thus, it is important that the validation of these bifurcations point candidates be made accurately in order to identify the true ones. It must be avoided the possibility of homogeneous bifurcations regions, where no real vessel bifurcation occurs, be validated. Those regions can introduce errors in the feature matching step.

The implemented characterization of the bifurcation points reveled to be quite sensible to the structure of the bifurcation region. Small deviations, from the ideal bifurcation region structure, produce wrong characterization of the bifurcation point and, thus, the wrong definition of the invariants and, consequently, introduce rough errors in the initial feature matching step.

Although the measures used to defined an initial set of matches between the features (mutual information and Euclidean distance in the space of invariants) were referred in other previous works, the criteria revealed to be too much basic (the pairs with maximal mutual information are matched approach due the pairs with the minimum Euclidean distance are matched). This caused that some features were considered the best initial match to many of the other image features. This caused the inliers selection steps to fail. The test of the steps of the inliers selection, transformation estimation and image re-sampling with manually matched features or with the registration of pairs of equal usage revealed that this implementation work properly.

## *6.2 Future work*

The implementation of the image registration algorithm in the visualization module of BW-Eye will be a valuable tool of that application. Although the described work did not reproduce the expected results, it can be a source of important analysis and useful conclusions. In future developments of this functionality of the application BW-Eye, it must be strictly defined the real requisites of that the algorithm must fulfil (type of images, number of images, the extension of the regions of the retina represented in the set of images, size of images, optimal duration of the registration process, accuracy) in order to select an approach that corresponds to the required requisites. As described during this document, there are a great variety of possible approaches to implement a registration algorithm, each one adapted and optimized to solve a particular image registration challenge with certain characteristics, in a particular domain. If the strictly analyses referred before is not made, it will be tried to implement an ideal image registration algorithm which will try to solve simultaneously all the challenges of the retinal image registration algorithm identified in this document, which is quite improbable.

# References

[1] A. Santos, "WIA-DM Ophtalmologic Decision Support System based on Clinical Workflow and Data Mining Techniques.", Project Course Report, University of Coimbra, 2007

[2] E. Ferreira, "WIA-DM Ophtalmologic Decision Support System based on Clinical Workflow and Data Mining Techniques.", Project Course Report, University of Coimbra, 2007

[3] P. Barbeiro, "WIA-DM Ophtalmologic Decision Support System based on Clinical Workflow and Data Mining Techniques.", Project Course Report, University of Coimbra, 2007

[4] T. Schlote, J. Rohrbach, M. Grueb and J.Mielke, *Pocket Atlas of Ophthalmology*. Georg Thieme Verlag, 2006.

[5] http://en.wikipedia.org/wiki/Eye   (accessed: August 2008)

[6] C. A. Bradford, *Basic ophthalmology for medical students and primary care residents.* American Academy of Ophthalmology, 1999.

[7] http://www.opsweb.org/OpPhoto/Angio/index.html   (accessed: August 2008)

[8] http://www.visionrx.com/library/enc/enc_angiography.asp (accessed: August 2008)

[9] http://en.wikipedia.org/wiki/Fundus_camera   (accessed: August 2008)

[10] A. A. Goshtasby, *2-D and 3-D Image Registration for Medical, Remote Sensing and Industrial Appications*. John Wiley & Sons, 2005

[11] I. N. Bankman, *Handbook of Medical Imaging Processing and Analysis.* Academic Press, 2000

[12] A. Can, C. Stewart, B. Roysam and H. Tanenbaum, "A Feature-Based, Robust, Hierarchical Algorithm for Registering Pairs of Images of the Human Curved Retina", *IEEE Trans. Pattern Analysis and Machine Intelligence*, vol. 24, no. 3, pp. 347-364, 2002.

[13] A. Can, C. Stewart, B. Roysam and H. Tanebaum, "A Feature-Based Technique for Joint, Linear Estimation of High-Order Image-to-Mosaic Transformations: Mosaicing the Curved Human Retina", *IEEE Trans. on Pattern Analysis and Machine Intelligence*, vol.24, no. 3, pp. 412-419, 2002.

[14] C. Stewart, "Computer Vision Algorithms for Retinal Image Analysis: Current Results and Future Directions", *Lecture Notes in Computer Science*, vol. 3765, pp. 31-50, 2005.

[15] T. Chanwimaluang, G. Fan and S. Fransen, "Hybrid Retinal Image Registration", *IEEE Trans. On Information Technology in Biomedicine*, vol. 10, no. 1, pp. 129-142, 2006.

[16] A.V. Cideciyan, "Registration of Ocular Fundus Images", *IEEE Eng. Med. Biol. Mag.*, vol. 14, pp. 52-58, Jan. 1995.

[17] N. Ritter, R. Owens, J. Cooper, R. Eikelboom and P. Van Saatloos, "Registration of Stereo and Temporal Images of the Retina", *IEEE Trans. Med. Imag.*, vol. 18, pp. 404-418, May 1999.

[18] M. Martinez-Perez, A. Hughes, S. Thom, A. Bharath, K. Parker, "Segmentation of Blood Vessels from Red-Free and Fluorescein Retinal Images", *Medical Image Analysis*, vol. 11, pp. 47-61, 2007

[19] A. Hoover, V. Kouznetsova and M. Goldbaum, "Locating Blood  Vessels in Retinal Images by Piecewise Threshold Probing of a Matched Filter Response", *IEEE Trans. Med. Imaging.*, vol. 19, no. 3, pp. 203-210, 2000.

[20] A. Hoover and M. Goldbaum, "Locating the Optic Nerve in a Retinal Image using the Fuzzy Convergence of the Blood Vessels", *IEEE Trans. Med. Imaging.*, vol. 22, no. 8, pp. 951-958, 2003.

[21] J. Lowell, A. Hunter, D. Steel, A. Basu, R. Ryder and E. Fletcher, " Optic Nerve Head Segmentation", *IEEE Trans. Med. Imaging.*, vol. 23, no. 2, pp. 256-264, 2004.

[22] A. Pinz, S. Bernogger, P. Datlinger and A. Kruger, "Mapping the Human Retina", *IEEE Trans. Med. Imaging.*, vol. 17, no. 4, pp. 606-620, Aug. 1998.

[23] A. Can, H. Shen, J. Turner, H. Tanenbaum and B. Roysam, "Rapid Automated Tracing and Feature Extraction from Retinal Fundus Images Using Direct Exploratory Algorithms", *IEEE Transactions on Information Technology in Biomedicine*, vol.3, no. 2, pp. 125-138, June 1999.

[24] T. Choe, I. Cohen, M. Lee, G. Medioni, "Optimal Global Mosaic Generation from Retinal Images", *Proceedings of the 18[th] International Conference on Pattern Recognition*, vol. 3, pp. 681-684, 2006.

[25] T. Choe, I. Cohen, "Registration of Multimodal Fluorescein Images Sequence of the Retina", *Proceedings of the 10th International Conference on Computer Vision*, vol. 1, pp. 106-113, 2005.

[26] P. Cattin, H. Bay, L. Gool and G. Székely, "Retina Mosaicing Using Local Features", *Lecture Notes in Computer Science*, vol. 4191, pp. 185-192, 2006.

[27] H. Bay, T. Tuytelaars, L. Gool, "SURF: Speeded Up Robust Features", *Lecture Notes in Computer Science*, vol. 3951, pp. 404-417, 2006.

[28] B. Zitová, J. Flusser, "Image Registration Methods: a Survey", *Image and Vision Computing*, vol. 21, pp. 977-1000, 2003.

[29] F. Zana, J. Klein, "A Multimodal Registration Algorithm of Eye Fundus Images Using Vessels Detection and Hough Transform", *IEEE Transactions on Medical Imaging*, vol.18, no. 5, pp. 419-428.

[30] Y. Bouaoune, M. Assogba, J. Nunes, P. Bunel, "Spatio-temporal Characterization of Vessel Segments Applied to Retinal Angiographic Images", *Pattern Recognition Letters*, vol. 24, pp. 607-615, 2003.

[31] C. Heneghan, P. Maguire, N. Ryan and P. Chazal, "Retinal Image Registration Using Control Points", *IEEE Inter. Symp. Biomedical Imaging,* pp. 349-352, July 2002.

[32] http://en.wikipedia.org/wiki/Information_entropy (accessed: August 2008)

[33] http://en.wikipedia.org/wiki/Mutual_information (accessed: August 2008)

[34] http://www.vcipl.okstate.edu/research_blood_vessel.html (accessed: August 2008)

[35] T. Chanwimaluang, G. Fan, S. Fransen, "Correction to Hybrid Retinal Image Registration", *IEEE Transactions on Information Technology in Biomedicine*, vol. 11, no. 1, pp.110, 2007.

[36] http://www.mathworks.com/matlabcentral/fileexchange/loadFile.do?objectId=4145&objectType=file (accessed: August 2008)

[37] http://en.wikipedia.org/wiki/Homography  (accessed: August 2008)

[38] http://en.wikipedia.org/wiki/RANSAC    (accessed: August 2008)

[39]
http://www.mathworks.com/matlabcentral/fileexchange/loadFile.do?objectId=18555&objectType
=file (accessed: August 2008)

# Appendices

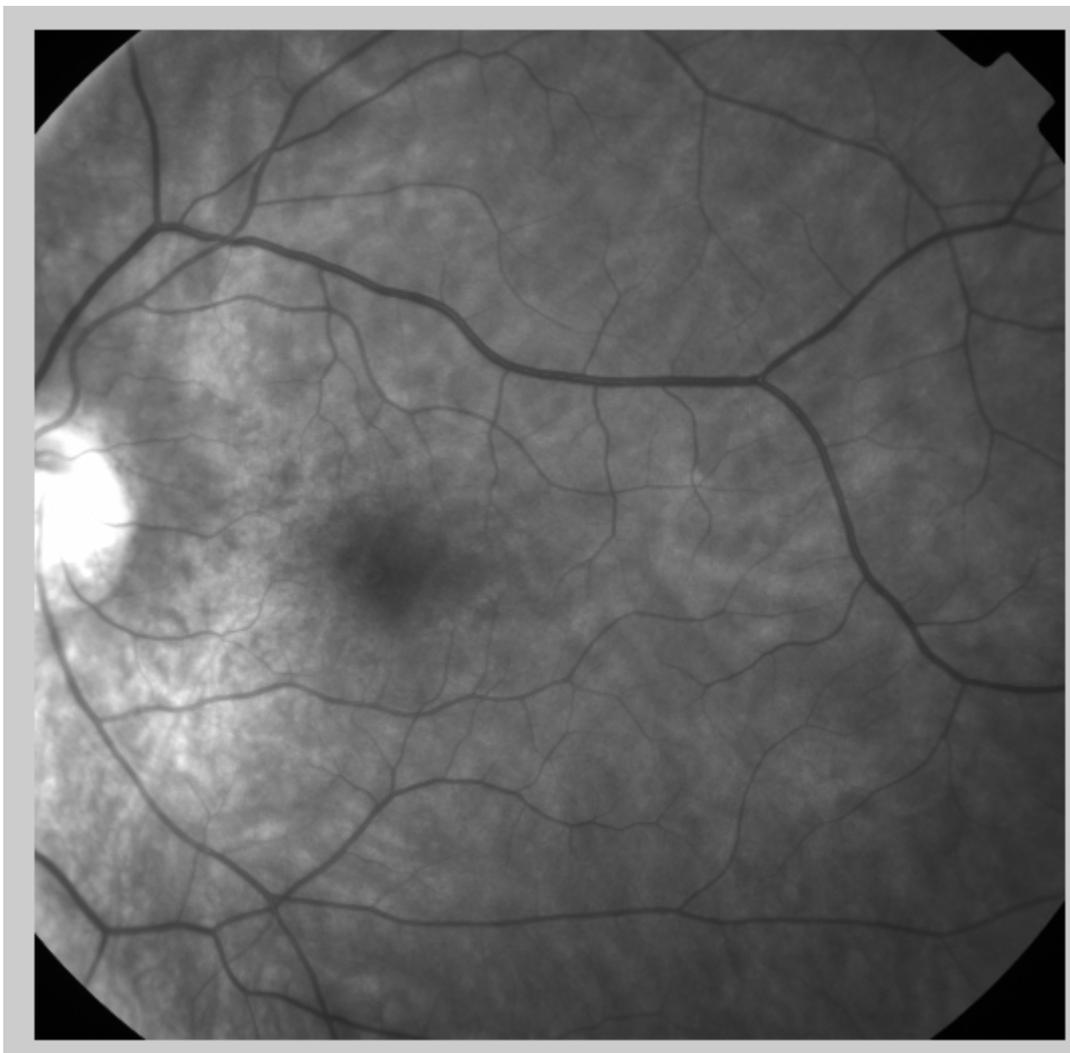

**Figure 1 – Original red-free retinography image.**

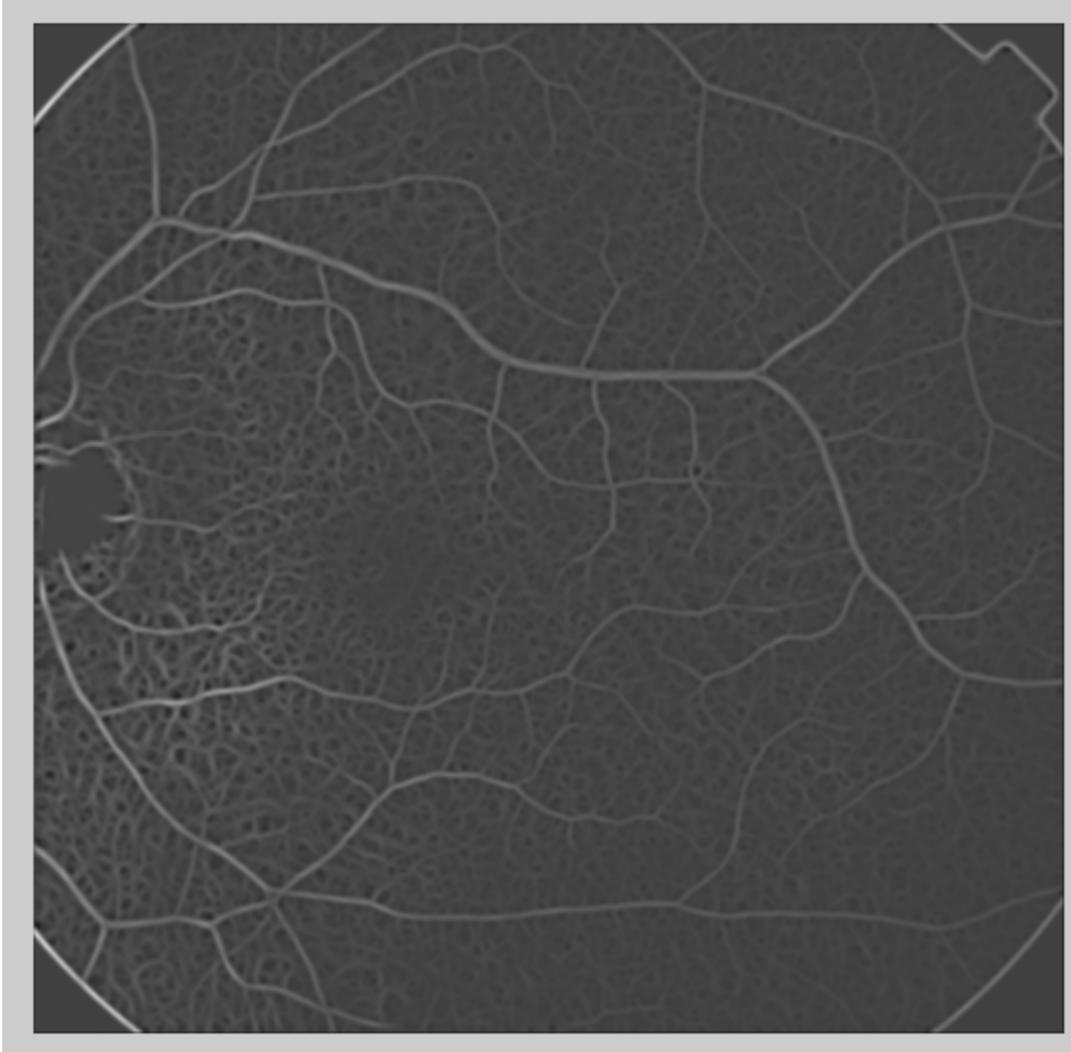

**Figure 2 – Pre-processed red-free retinography image**

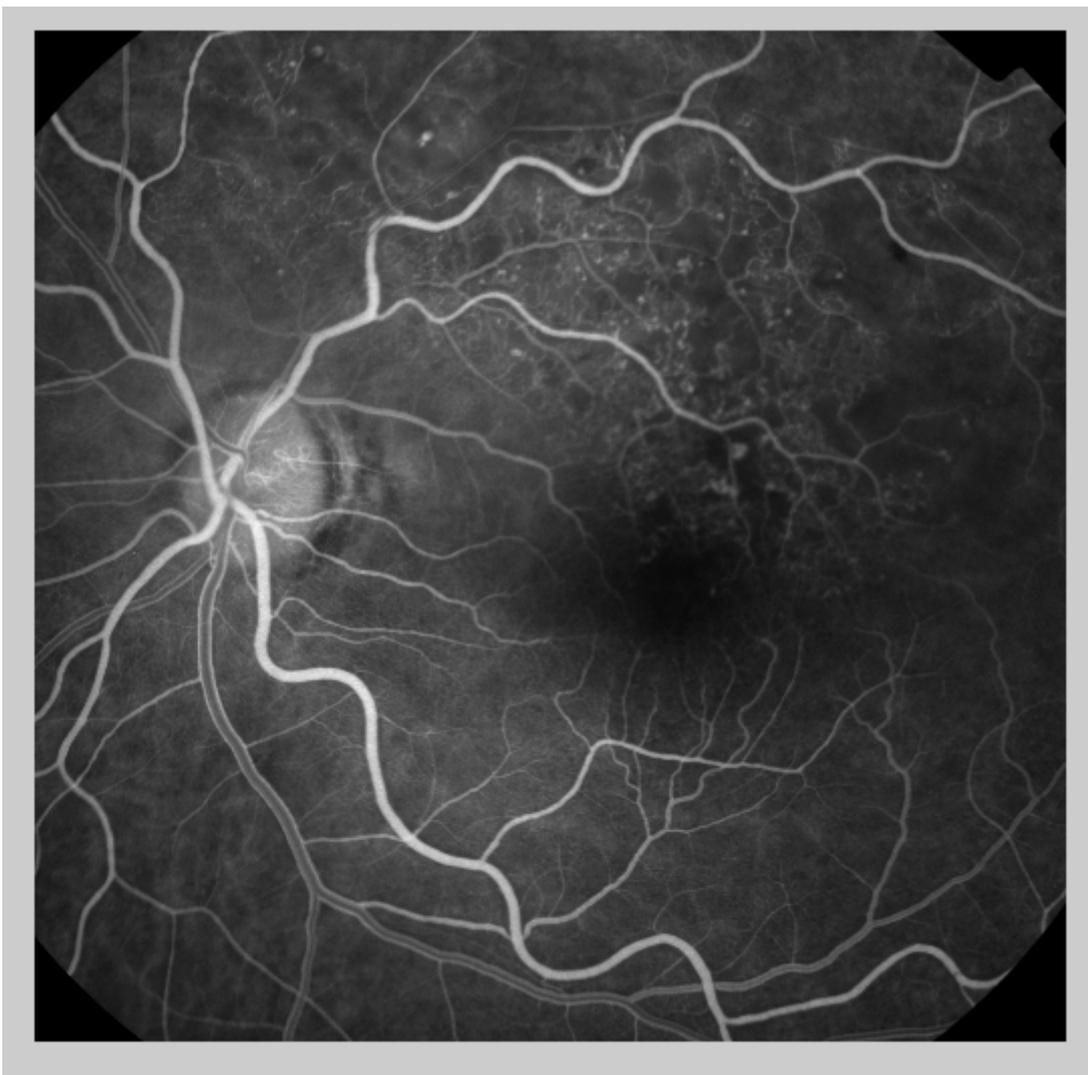

**Figure 3** – Original retinal angiography image.

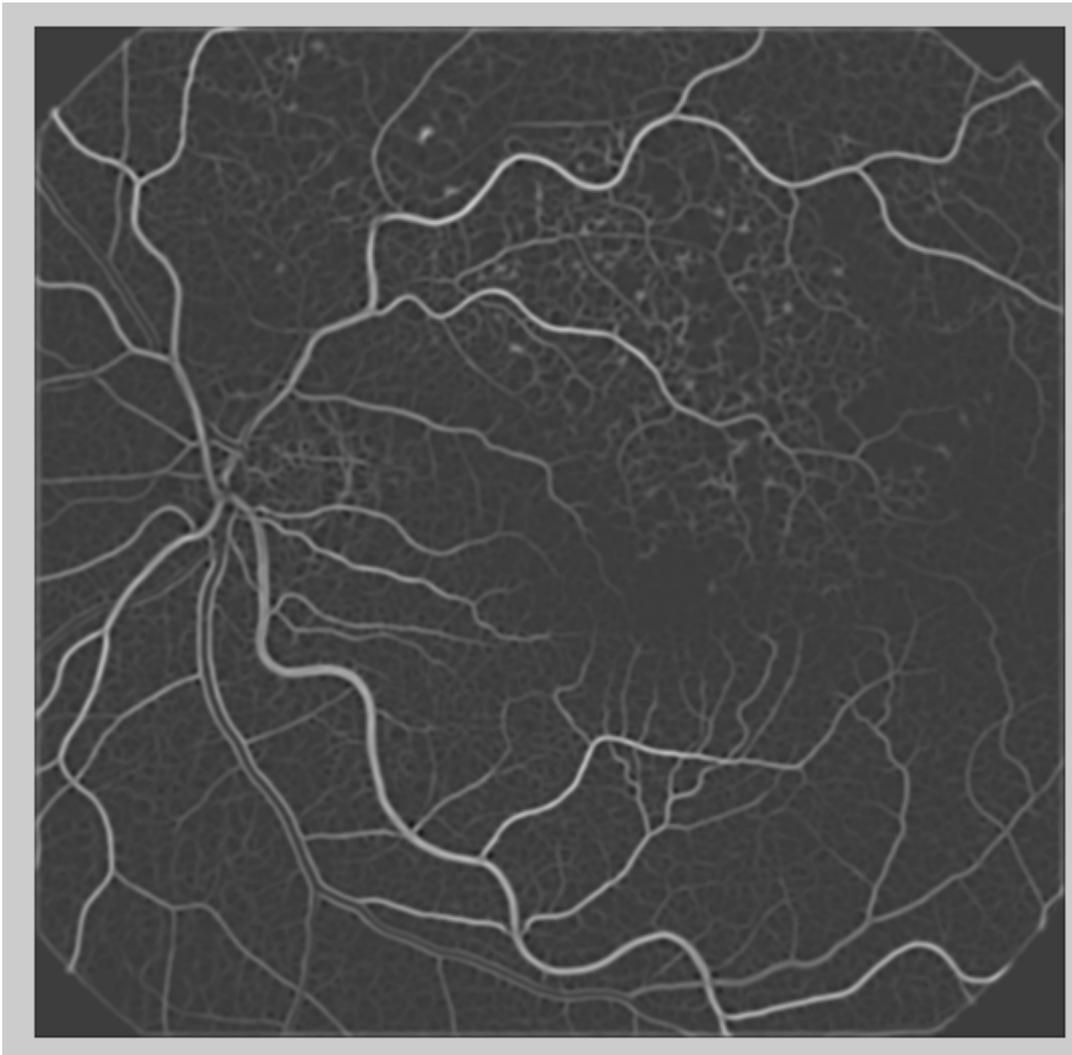

**Figure 4 – Pre-processed retinal angiography image.**

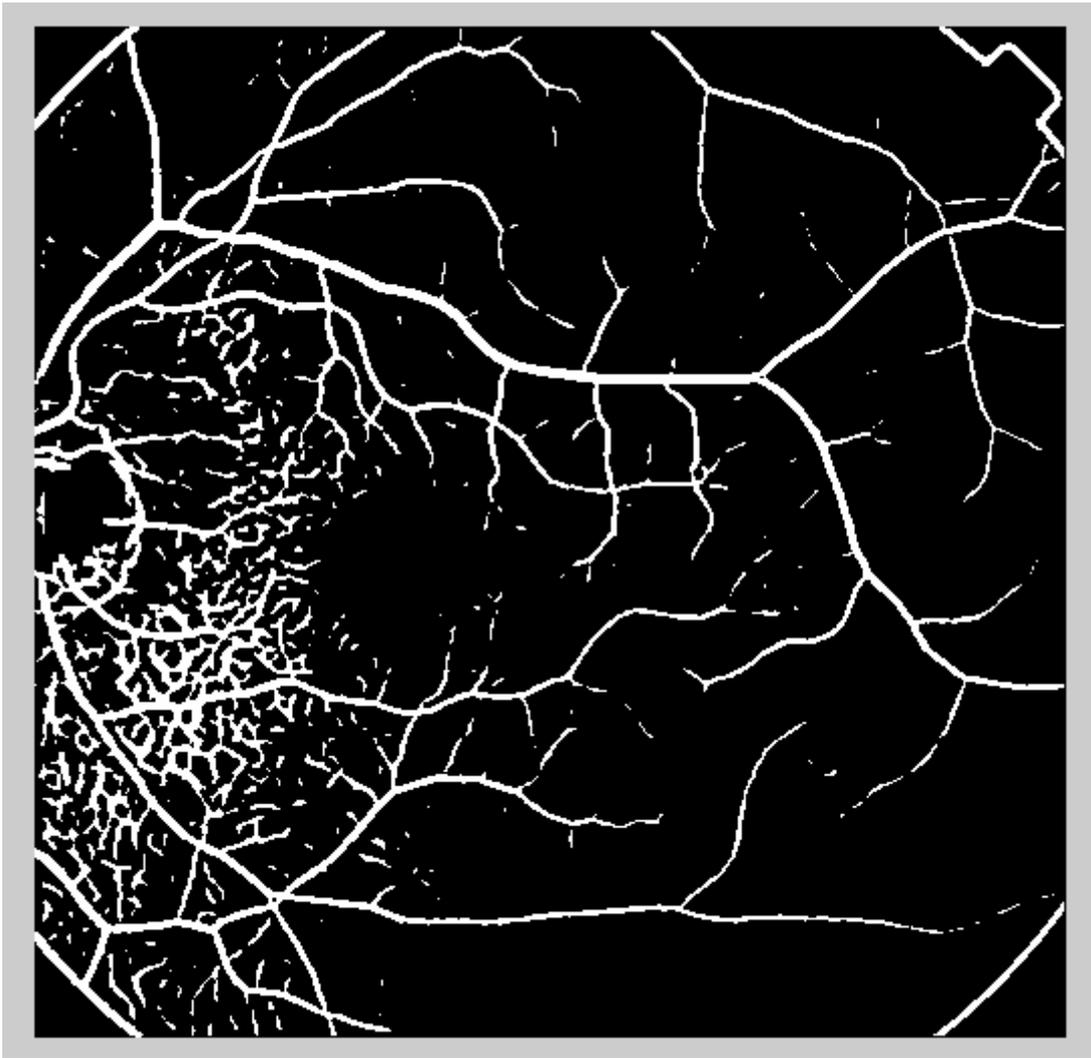

**Figure 5 – Segmented red-free retinography image**

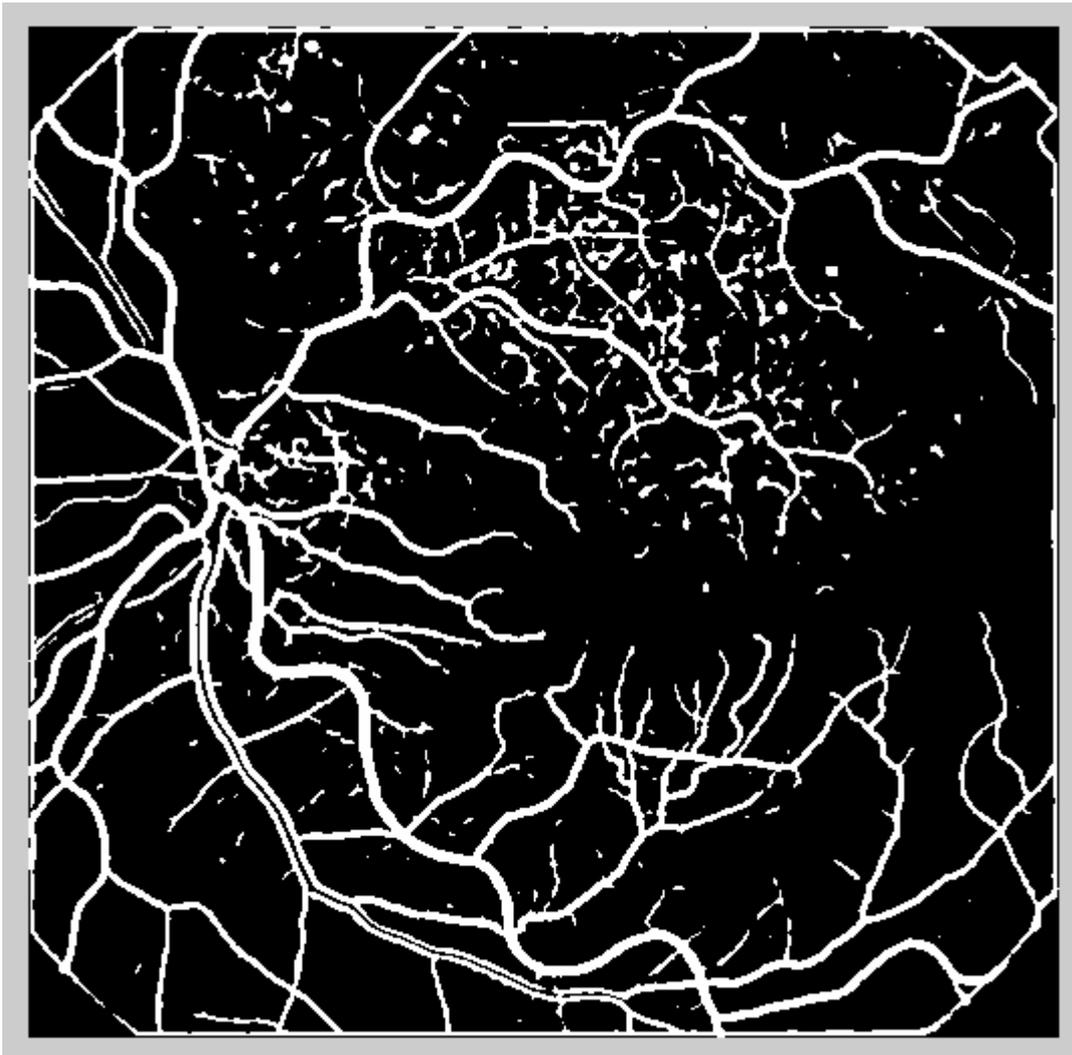

**Figure 6 – Segmented retinal angiography image.**

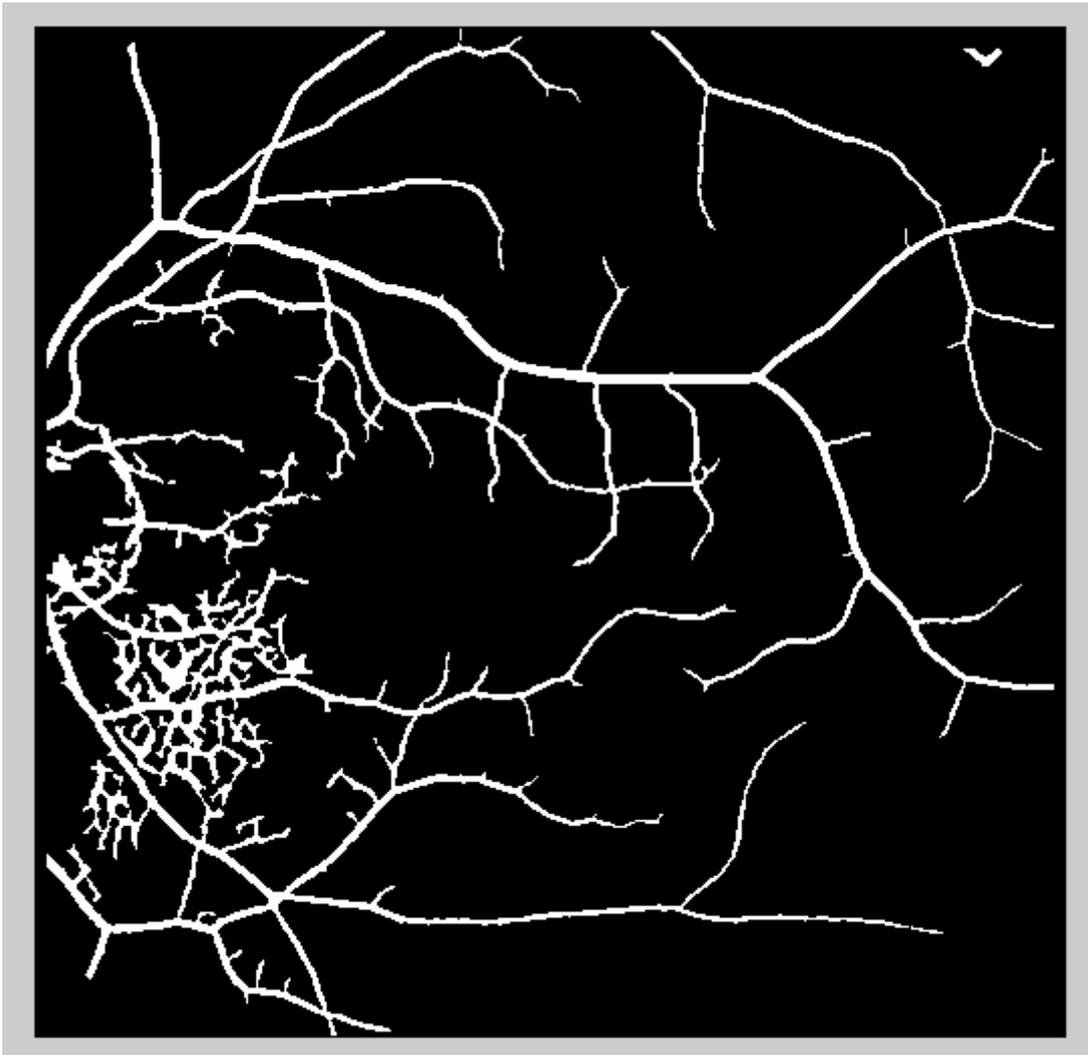

**Figure 7 -  Segmented red-free retinography image after the application of the size filtering, hollow vessels filling and the mask remotion.**

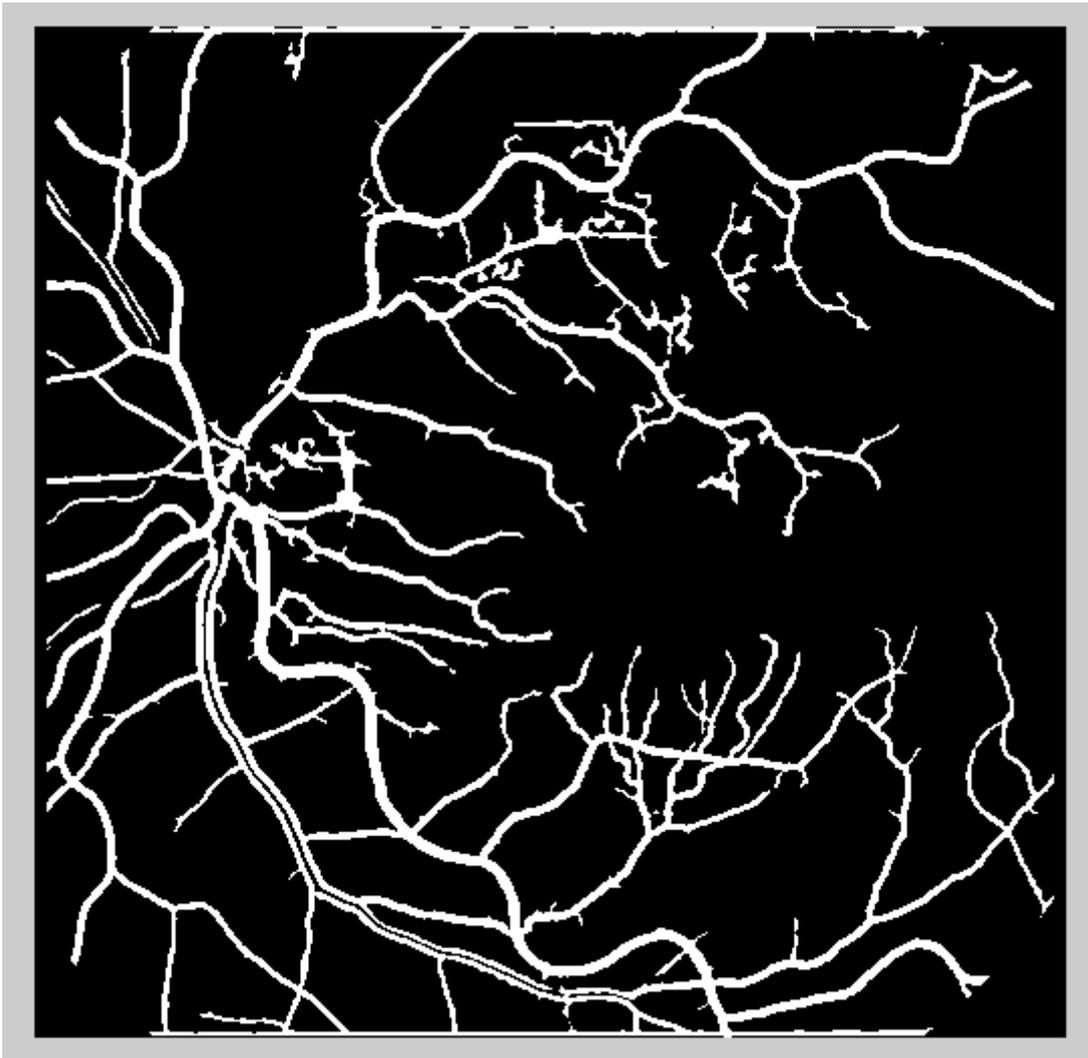

**Figure 8 - Segmented retinal angiography image after the application of the size filtering, hollow vessels filling and the mask remotion.**

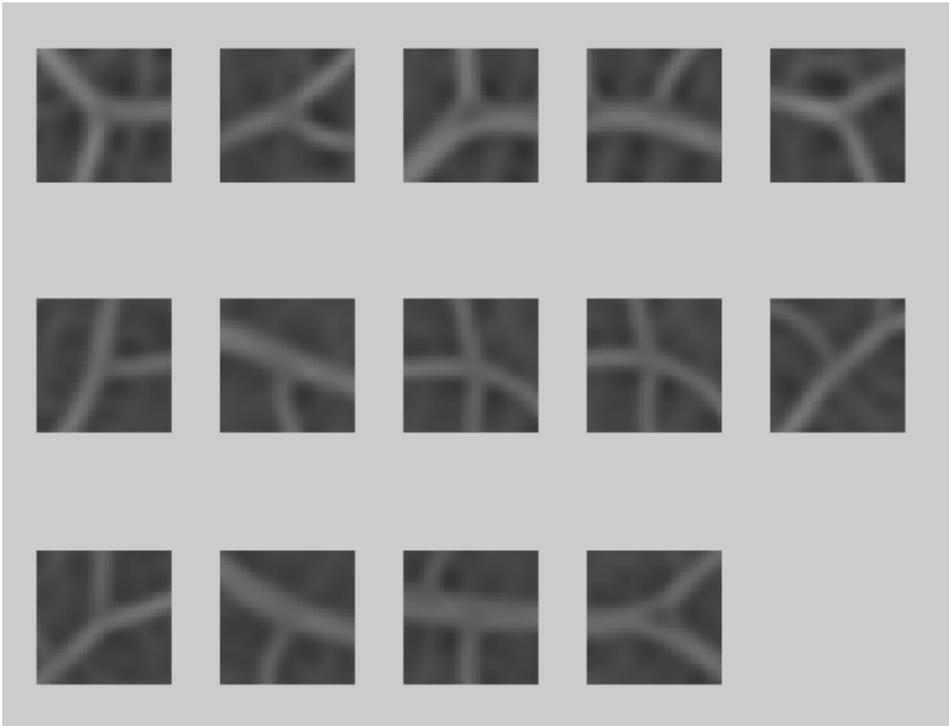

**Figure 9 – Bifurcation regions of the validated bifurcation points of a red-free retinography image.**

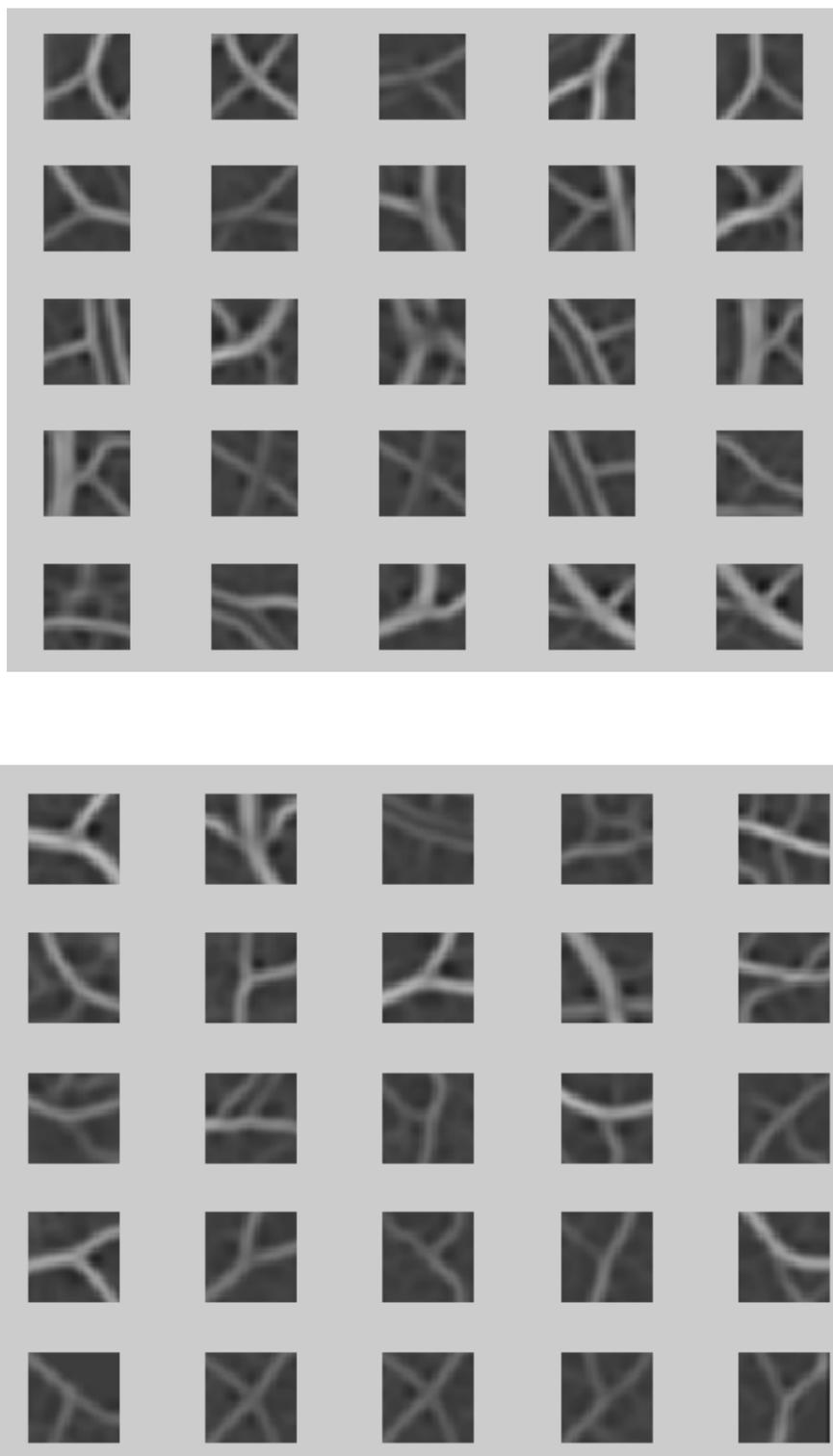

**Figure 10 – Bifurcation regions of the validated bifurcation points of a retinal angiography image.**

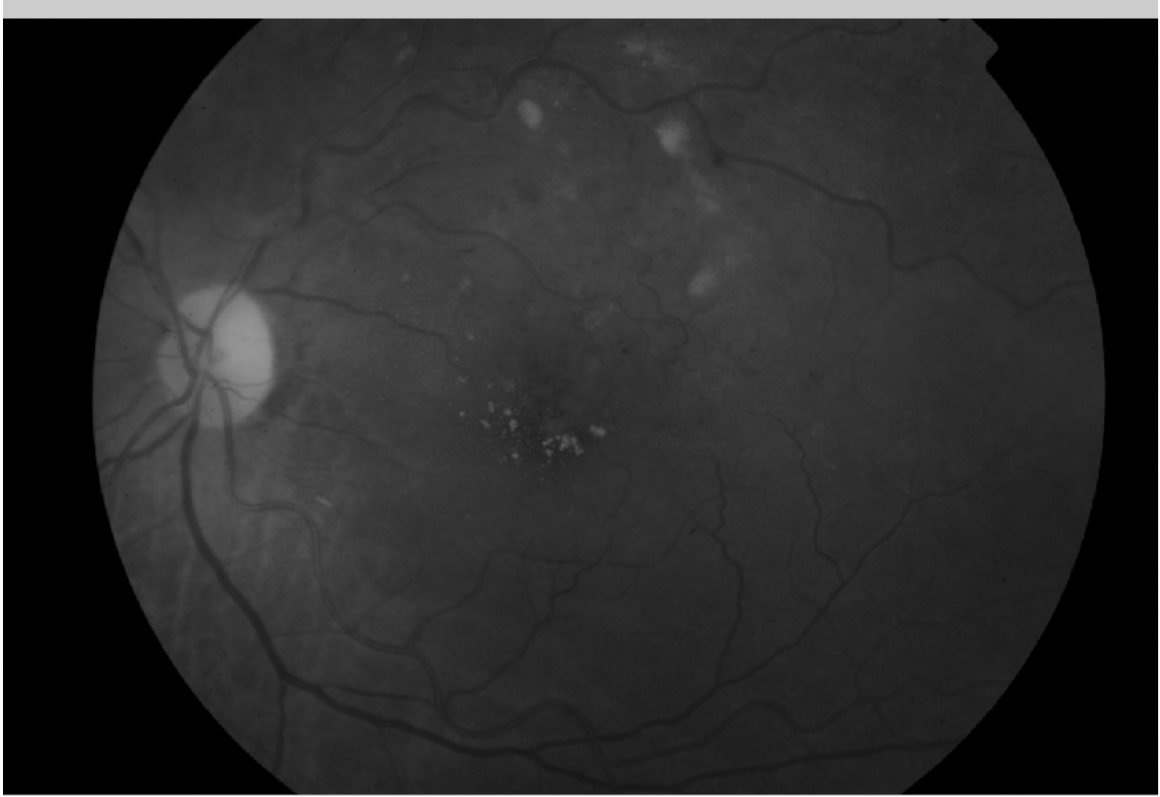

**Figure 11 –Original red-free image used in the tests of the inliers selection step**

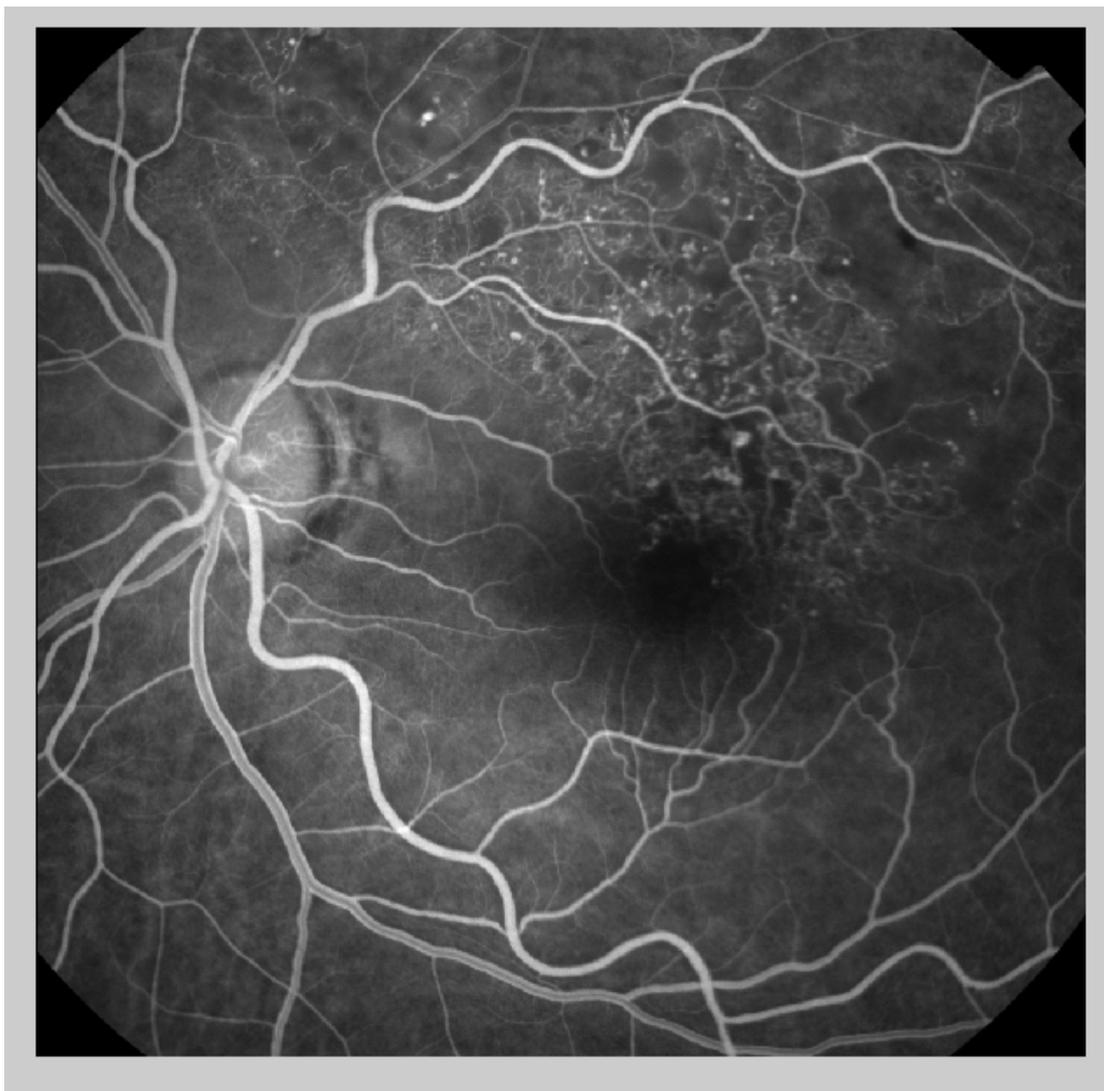

**Figure 12–Original retinal angiography image used in the tests of the inliers selection step**

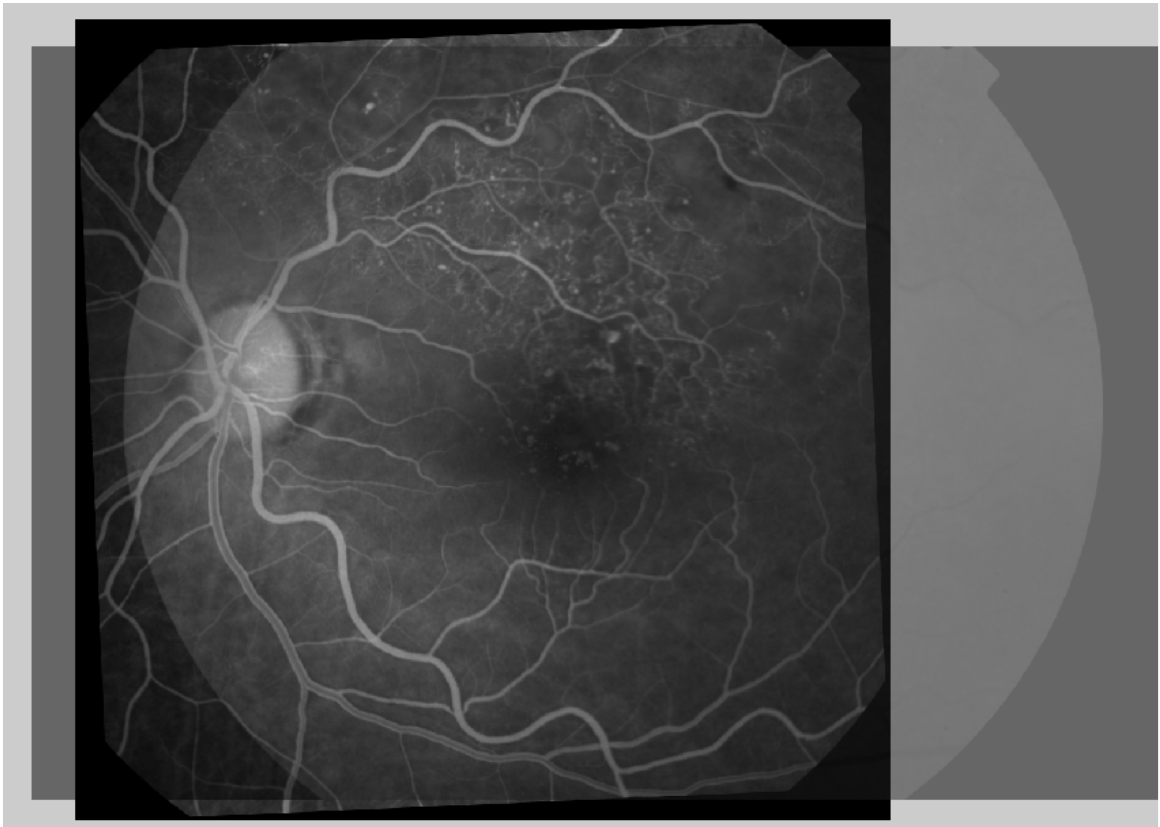

**Figure 13 – Registered pair of images that results from the test of the inliers selection step with a transparence of 50%.**